\pdfoutput=1

\documentclass[10pt,twocolumn,letterpaper]{article}

\usepackage[pagenumbers]{cvpr} 

\usepackage{graphicx}
\usepackage{amsmath}
\usepackage{amssymb}
\usepackage{booktabs}
\usepackage{enumitem}

\usepackage[accsupp]{axessibility}  

\usepackage{comment}

\usepackage[pagebackref,breaklinks,colorlinks]{hyperref}

\usepackage[export]{adjustbox}

\usepackage[capitalize]{cleveref}
\crefname{section}{Sec.}{Secs.}
\Crefname{section}{Section}{Sections}
\Crefname{table}{Table}{Tables}
\crefname{table}{Tab.}{Tabs.}
\graphicspath{{images/}}

\usepackage{dsfont}
\usepackage{etoolbox}
\usepackage{color}

\newif\ifshowedits

\newcommand{\addeditor}[3]{%
  \definecolor{#1color}{rgb}{#3}
  \expandafter\newcommand\csname #1\endcsname[1]{%
  \ifshowedits
    {\color{#1color} ##1}%
  \else
    {##1}%
  \fi
  }%
  \expandafter\newcommand\csname #1rmk\endcsname[1]{%
  \ifshowedits
    {\color{#1color} {\bf [#2: ##1]}}
  \fi
  }%
  \expandafter\newcommand\csname #1rpl\endcsname[2]{%
  \ifshowedits
    {\color{#1color} ##1 \sout{##2}}
  \else
    {##1}
  \fi
  }%
}

\newcommand{\createtextvar}[1]{
  \expandafter\newcommand\csname #1\endcsname{%
  {\text{#1}}
}%
}
\newcommand{\textvars}[1]{\forcsvlist{\createtextvar}{#1}}

\newcommand{\mycomment}[1]{}

\newcommand{\calC}{{\cal C}}

\newcommand{\IR}{{\mathds{R}}}

\DeclareMathOperator*{\argmax}{arg\,max}

\addeditor{vincent}{VL}{0.0, 0.5, 0.0}
\addeditor{antoine}{AG}{0.0, 0.0, 0.8}
\addeditor{tom}{TM}{0.9, 0.5, 0.0}
\addeditor{todo}{TODO}{1.0, 0.0, 0.0}
\showeditsfalse

\textvars{pos,rot}

\def\cvprPaperID{10995}
\def\confName{CVPR}
\def\confYear{2023}

\begin{document}

\title{MACARONS: Mapping And Coverage Anticipation\\ with RGB Online Self-Supervision}

\author{Antoine Guédon \qquad Tom Monnier \qquad Pascal Monasse \qquad Vincent Lepetit\\
LIGM, Ecole des Ponts, Univ Gustave Eiffel, CNRS, France\\
{\tt\small \url{https://imagine.enpc.fr/~guedona/MACARONS/}}
}

\twocolumn[\maketitle\vspace{-3em}

\begin{center}
    \captionsetup{type=figure}
    \begin{subfigure}[t]{0.48\linewidth}
        \includegraphics[width=\linewidth]{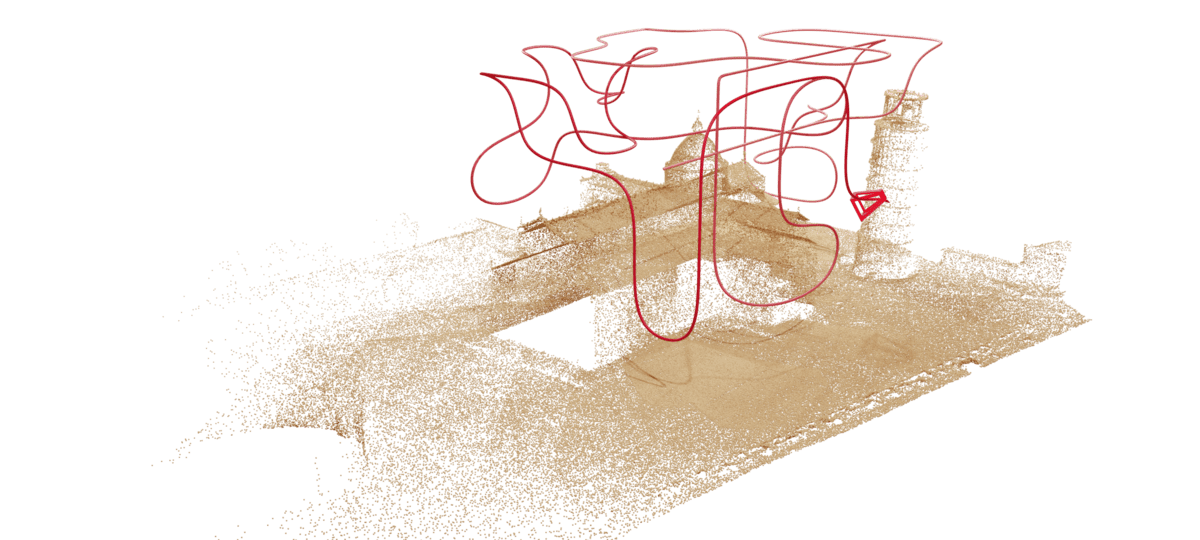}%
        \vspace{.5em}
        \includegraphics[width=\linewidth]{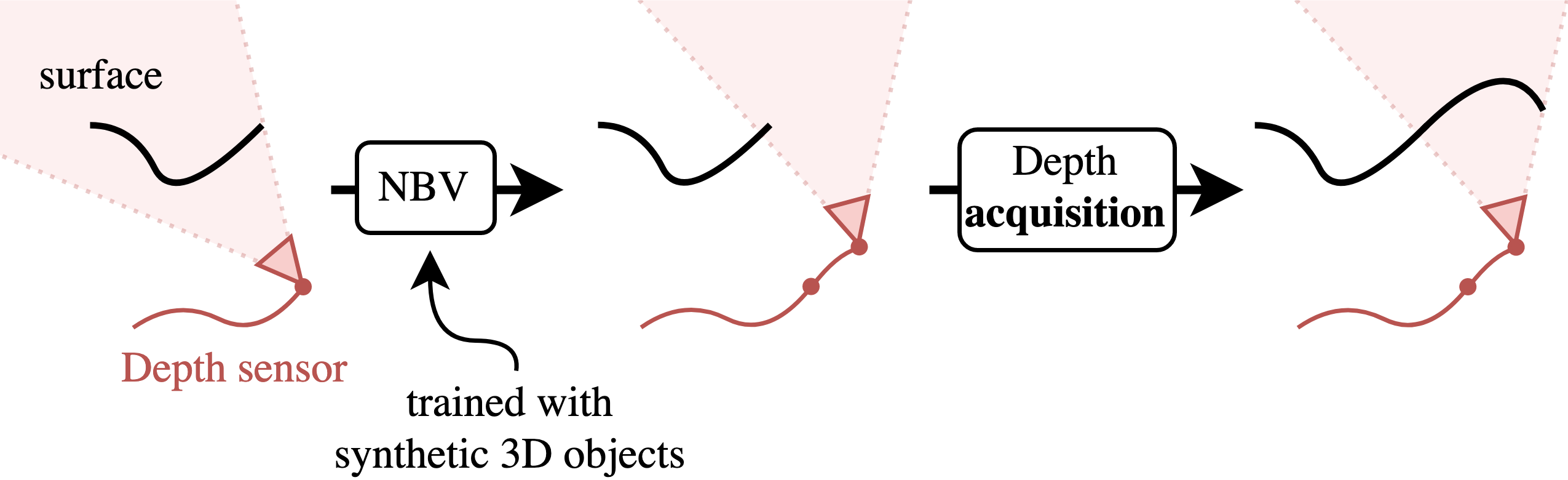}%
        \vspace{.1em}
        \caption{NBV methods with a depth sensor (\eg,~\cite{guedon-nips22-scone-surface-coverage})}
    \end{subfigure}
    \hfill
    \begin{subfigure}[t]{0.48\linewidth}
        \includegraphics[width=\linewidth]{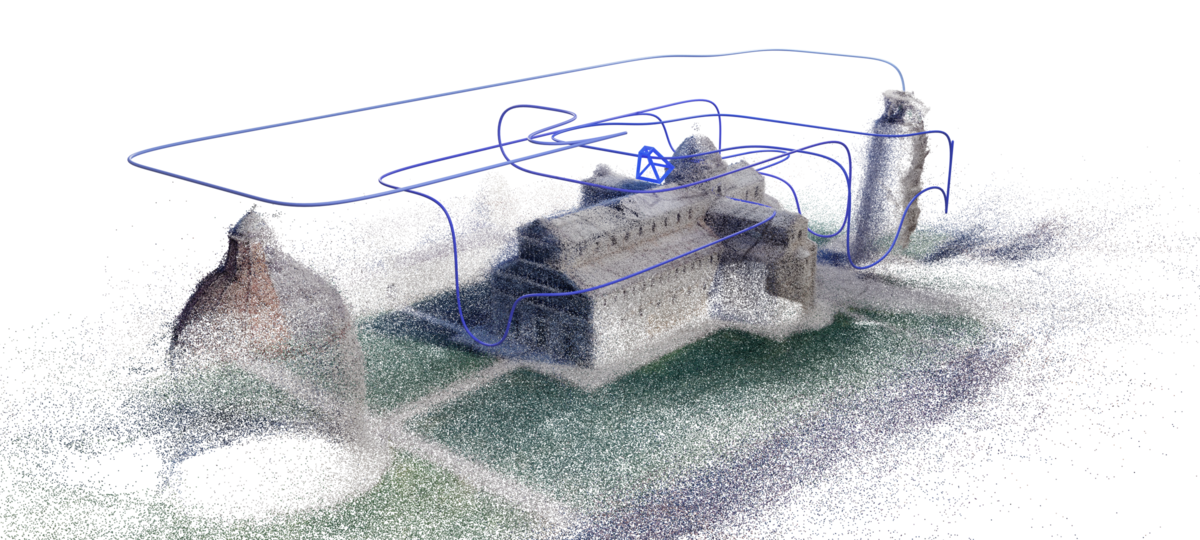}%
        \vspace{.5em}
        \includegraphics[width=\linewidth]{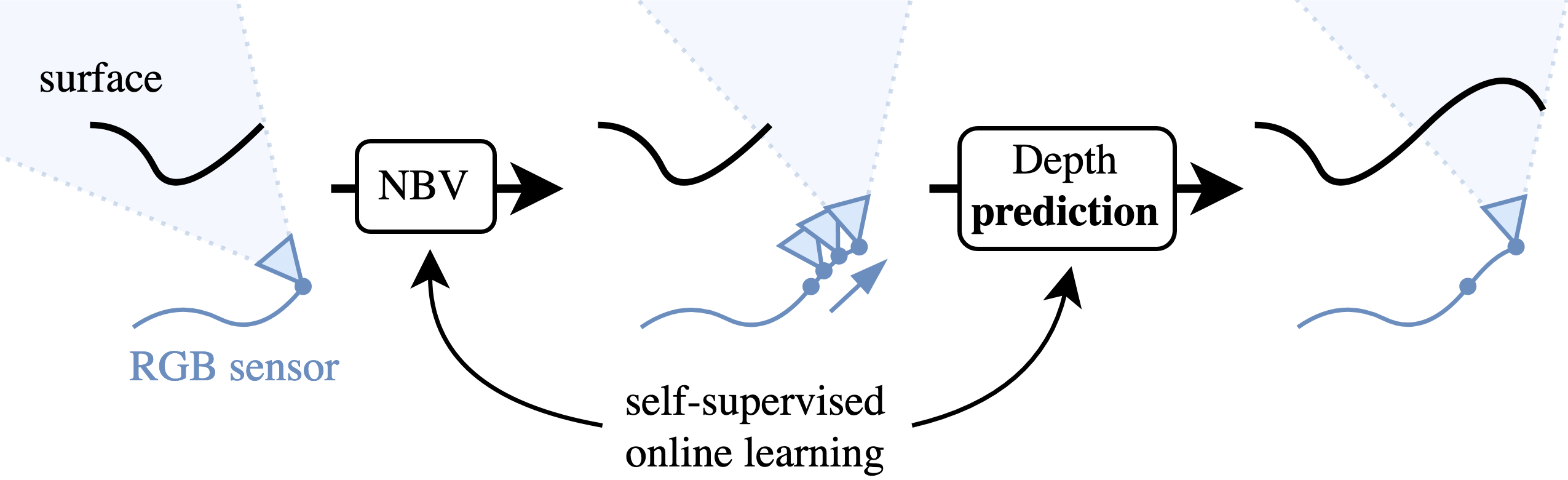}%
        \vspace{.1em}
        \caption{Our approach MACARONS with an RGB sensor}
    \end{subfigure}
\end{center}

\vspace{-1.5em}
\addtocounter{figure}{-1}
\captionof{figure}{\textbf{Mapping and Coverage Anticipation with RGB Online Self-Supervision. (a)} NBV methods such as \cite{guedon-nips22-scone-surface-coverage} rely on a depth sensor to perform path planning (bottom) and scan the environment (top). They need to be trained with explicit 3D supervision, generally on small-scale meshes. \textbf{(b)} Our approach MACARONS instead simultaneously learns to efficiently explore the scene and to reconstruct it (top) using an RGB sensor only. Its self-supervised, online learning process scales to large-scale and complex scenes.}
\label{fig:teaser}
\bigbreak]


\begin{abstract}
We introduce a method that simultaneously learns to explore new large environments and to reconstruct them in 3D from color images only. This is closely related to the Next Best View problem (NBV), where one has to identify where to move the camera next to improve the coverage of an unknown scene. However, most of the current NBV methods rely on depth sensors, need 3D supervision and/or do not scale to large scenes. Our method requires only a color camera and no 3D supervision. It simultaneously learns in a self-supervised fashion to predict a ``volume occupancy field'' from color images and, from this field, to predict the NBV. Thanks to this approach, our method performs well on new scenes as it is not biased towards any training 3D data. We demonstrate this on a recent dataset made of various 3D scenes and show it performs even better than recent methods requiring a depth sensor, which is not a realistic assumption for outdoor scenes captured with a flying drone.
\end{abstract}

\section{Introduction}
\label{sec:intro}

By bringing together Unmanned Aerial Vehicles~(UAVs) and Structure-from-Motion algorithms, it is now possible to reconstruct 3D models of large outdoor scenes, for example for creating a Digital Twin of the scene. However, flying a UAV requires expertise, especially when capturing images with the goal of running a 3D reconstruction algorithm, as the UAV needs to capture images that together cover the entire scene from multiple points of view. Our goal with this paper is to make this capture automatic by developing a method that controls a UAV and ensures a coverage suitable to 3D reconstruction.

This is often referenced in the literature as the ``Next Best View'' problem~(NBV)~\cite{connolly-father-paper}:  Given a set of already-captured images of a scene or an object, how should we move the camera to improve our coverage of the scene or object? Unfortunately, current NBV algorithms are still not suitable for three main reasons. First, most of them rely on a voxel-based representation and do not scale well with the size of the scene. Second, they also rely on a depth sensor, which is in practice not possible to use on a small UAV in outdoor conditions as it is too heavy and requires too much power. Simply replacing the depth sensor by a monocular depth prediction method~\cite{Ranftl2022, Wei2021CVPR, Miangoleh2021Boosting} would not work as such methods can predict depth only up to a scale factor. The third limitation is that they require 3D models for learning to predict how much a pose will increase the scene coverage.

In this paper, we show that it is possible to simultaneously learn in a self-supervised fashion to efficiently explore a 3D scene and to reconstruct it using an RGB sensor only, without any 3D supervision. This makes it convenient for applications in real scenarios with large outdoor scenes.  We only assume the camera poses to be known, as done in past works on NBV~\cite{zeng-icirs20-pcnbv, hepp-corr18-learntoscore, mendoza-prl2020}. This is reasonable as NBV methods control the camera.

The closest work to ours is probably the recent \cite{guedon-nips22-scone-surface-coverage}. \cite{guedon-nips22-scone-surface-coverage} proposed an approach that can scale to large scenes thanks to a Transformer-based architecture that predicts the visibility of 3D points from any viewpoint, rather than relying on an explicit representation of the scene such as voxels. However, this method still uses a depth sensor. It also uses 3D meshes for training the prediction of scene coverage. To solve this, \cite{guedon-nips22-scone-surface-coverage} relies on meshes from ShapeNet~\cite{chang-15-shapenet}, which is suboptimal when exploring large outdoor scenes, as our experiments show. This limitation can actually be seen in Figure~\ref{fig:teaser}: The trajectory recovered by \cite{guedon-nips22-scone-surface-coverage} mostly focuses on the main building and does not explore the rest of the scene. By contrast, we use a simple color sensor and do not need any 3D supervision.

As our experiments show, we nonetheless significantly outperform this method thanks to our architecture and joint learning strategy. As shown in Figure~\ref{fig:pipeline}, our architecture is made of three neural modules that communicate together:
\begin{enumerate}[align=left,labelsep=0cm,itemindent=-1mm,noitemsep]
\item Our first module learns to predict depth maps from a sequence of images in a self-supervised fashion. 
\item Our second module predicts a ``volume occupancy field'' from a partial surface point cloud. This field is made of the probability for any input 3D point to be occupied or empty, given the previous observed images of the scene. We train this module from past experience, with partial surface point cloud as input and aiming to predict the occupancy field computed from the final point cloud.
\item Our third module predicts for an input camera pose the ``surface coverage gain'', \ie, how much new surface will be visible from this pose. We improve the coverage estimation model introduced by \cite{guedon-nips22-scone-surface-coverage} and propose a novel, much simpler loss that yields better performance. We rely on this module to identify the NBV.
\end{enumerate}

While exploring a new scene and training our architecture, we repeat the three following steps:
(1) We identify the Next Best View where to move the camera; 
(2) We move the camera to this Next Best View, collect images along the way, and build a self-supervision signal from the collected images, which we store in the ``Memory'';
\vincent{(3) We update the weights of all 3 modules using Memory Replay~\cite{Mnih13}. This avoids catastrophic forgetting and significantly speeds up training compared to a training procedure that uses only recent data, as such data is highly correlated. This last step establishes a synergy between the different parts of the model, each one providing inputs to the other parts.}

We compare to recent work~\cite{guedon-nips22-scone-surface-coverage} on their dataset made of large scale 3D scenes under the CC license. We evaluate the evolution of total surface coverage by a sensor exploring several 3D scenes. Our online, self-supervised approach that learns from RGB images is able to have better results than state-of-the-art methods with a perfect depth sensor.

To summarize, we propose the first deep-learning-based NBV approach for dense reconstruction of large 3D scenes from RGB images. We call this approach MACARONS, for Mapping And Coverage Anticipation with RGB Online Self-Supervision. Moreover, we provide a dedicated training procedure for online learning for scene mapping and automated exploration based on coverage optimization in any kind of environment, with no explicit 3D supervision. Consequently, our approach is also the first NBV method to learn in real-time to reconstruct and explore arbitrarily large scenes in a self-supervised fashion. We experimentally show that this greatly improves results for NBV exploration of 3D scenes. It makes our approach suitable for real-life applications on small drones with a simple color camera. More fundamentally, it shows that an autonomous system can learn to explore and reconstruct environments without any 3D information \emph{a priori}. We will make our code available on a dedicated webpage for allowing comparison with future methods.
\section{Related Work}
\label{sec:relatedwork}

We first review prior works for next best view computation. We then discuss depth estimation literature, from which we borrow techniques to avoid the need for depth acquisition.

\subsection{Next Best View (NBV)}
Approaches to NBV can be broadly split into two groups based on the scene representation. On the one hand, volumetric methods represent the scene as voxels used as inputs of traditional optimization schemes~\cite{potthast-14-probabilistic-framework, vasquez-14-volumetric-nbv, charrow-15-information, bissmarck-15-efficient-nbv, vasquez-17-view-state-planning, daudelin-17-adaptable-probabilistic, cieslewski-17-ICIRS, wang-20-efficient-autonomous} or more recently, neural networks~\cite{hepp-corr18-learntoscore, mendoza-prl2020, vasquez-21-3DCNN}. 
On the other hand, surface-based approaches~\cite{chen-smc05-visionsensorplanning, kriegel-11-surface-based, kriegel-12-next-best-scan, lee-tra20-mechanical-parts, zeng-icirs20-pcnbv} operate on dense point clouds representing the surface as computed by the depth sensor. Although modeling surfaces allows to preserve highly-detailed geometries, it does not scale well to complex scenes involving large point clouds, thus limiting their applicability to synthetic settings of isolated centered objects with cameras lying on a sphere. The closest work to ours is Guédon~\etal~\cite{guedon-nips22-scone-surface-coverage} which proposes an hybrid approach called SCONE that maximizes the surface coverage gain using a volumetric representation. Our proposed approach yet differs in two ways. First, although SCONE processes real complex scenes with free camera motions at inference, it can only be trained on synthetic datasets~\cite{shapenet2015}. Our approach instead benefits from a new online self-supervised learning strategy, which is the source of our better performances. Second, like most of NBV methods, SCONE assumes to have access to a depth sensor whereas our framework relies on RGB images only.

To relax the need for depth acquisitions, we propose a self-supervised method that learns to predict a depth map from color images captured by an arbitrary RGB sensor such as a flying drone while exploring a new environment.

\subsection{Depth estimation}

\paragraph{Monocular.} Classical monocular deep estimation methods are learned with explicit supervision, using either dense annotations acquired from depth sensors~\cite{eigen2014depth,eigen2015predicting,fu2018deep} or sparse ones from human labeling~\cite{chen2016single}. Recently, other works used self-supervision to train their system in the form of reprojection errors computed using image pairs~\cite{garg2016unsupervised,xie2016deep3d, godard2017unsupervised} or frames from videos~\cite{zhou2017unsupervised,gordon2019depth,zhao2022monovit}. Advanced methods even incorporate a model for moving objects~\cite{ranjan2018adversarial,godard2019digging,gordon2019depth,godard-iccv19-digginginto,geonet2018,chen2019self,bian2019unsupervised, li2019mannequin, tosi2020distilled, klingner2020self,li2020unsupervised}. However, all these approaches are typically self-trained and evaluated on images from a specific domain, whereas our goal is to obtain robust performances for any environment and any RGB sensor.

\paragraph{Sequential monocular.} A way to obtain better depth predictions during inference is to assume the additional access to a sequence of neighboring images, which is the case in our problem setup. Traditional non-deep approaches are efficient methods developed for SLAM~\cite{newcombe2011dtam,newcombe2010live,engel2014lsd,yang2018polarimetric}, which can further be augmented with neural networks~\cite{tateno2017cnn,bloesch2018codeslam,laidlow2019deepfusion}. Deep approaches typically refine at test time monocular depth estimation networks to account for the image sequence~\cite{casser2018depth,chen2019self,luo2020consistent,mccraith2020monocular, shu2020feature,kuznietsov2021comoda}. Other methods instead modify the architecture of monocular networks with recurrent layers to train directly with sequences of images~\cite{kumar2018depthnet, zhang2019exploiting, xie2019video, wang2019recurrent, patil2020dont, wang2020self}.
Inspired by deep stereo matching approaches~\cite{vzbontar2016stereo,luo2016efficient,shaked2017improved,kendall2017end,chang2018pyramid,Zhang2019GANet,cheng2019learning,zhang2019domaininvariant}, another line of works utilizes 3D cost volumes to reason about the underlying geometry at inference~\cite{liu2019neural, hou2019multi, wu2019spatial, wimbauer2020monorec, watson-cvpr21-temporal-opportunist-manydepth, guizilini-cvpr2022-multiframe}. In particular, the work of Watson~\etal~\cite{watson-cvpr21-temporal-opportunist-manydepth} introduces an efficient cost volume based depth estimator that is self-supervised from raw monocular videos and that provides state-of-the-art results. In this work, we adapt the self-supervised learning framework from~\cite{watson-cvpr21-temporal-opportunist-manydepth} to jointly learn our NBV and depth estimation modules.
\section{Problem setup}
\label{sec:problem_setup}
\begin{figure*}
  \centering
    \includegraphics[width=0.54\linewidth]{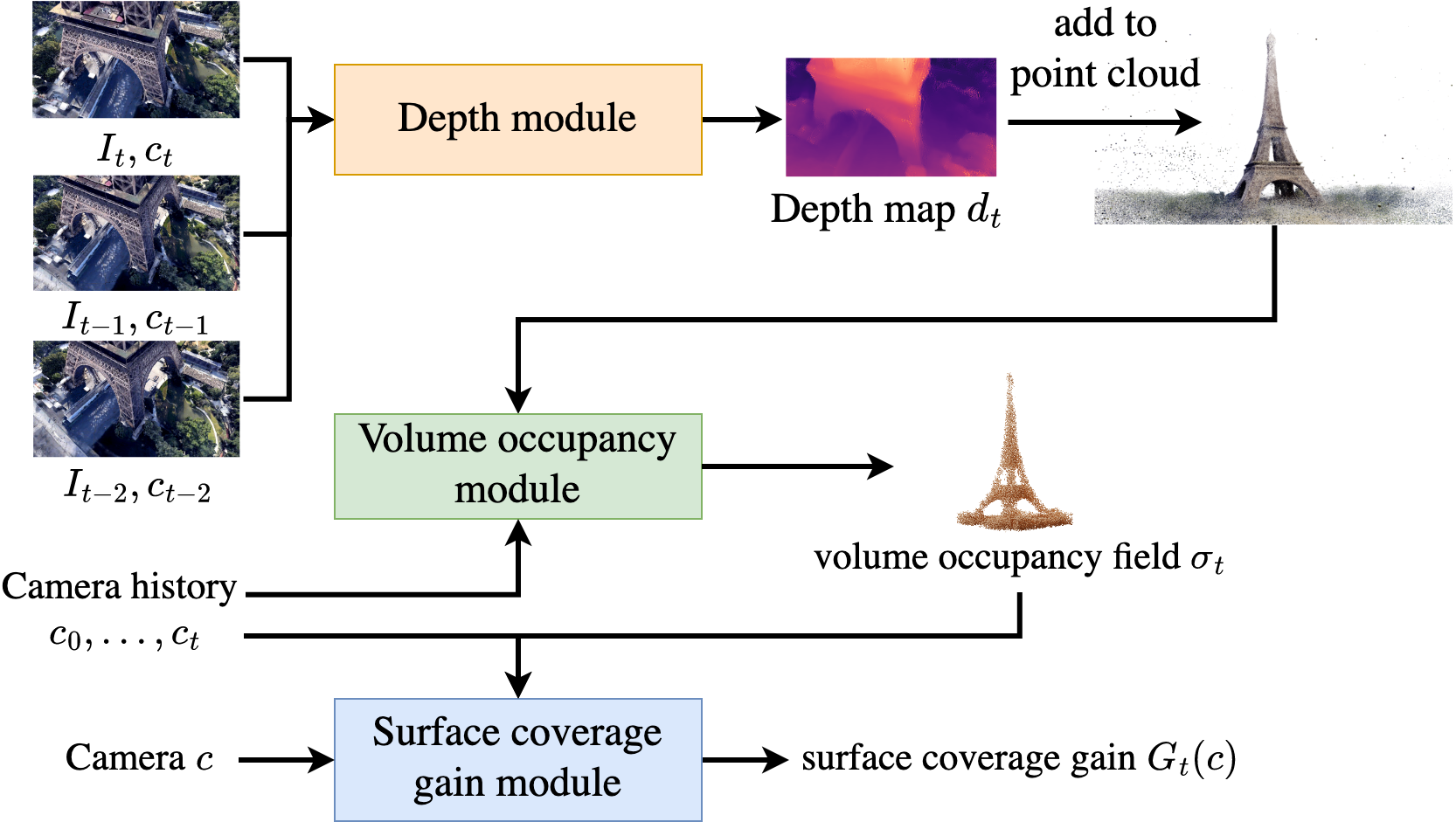}%
    \hfill
    \includegraphics[width=0.42\linewidth]{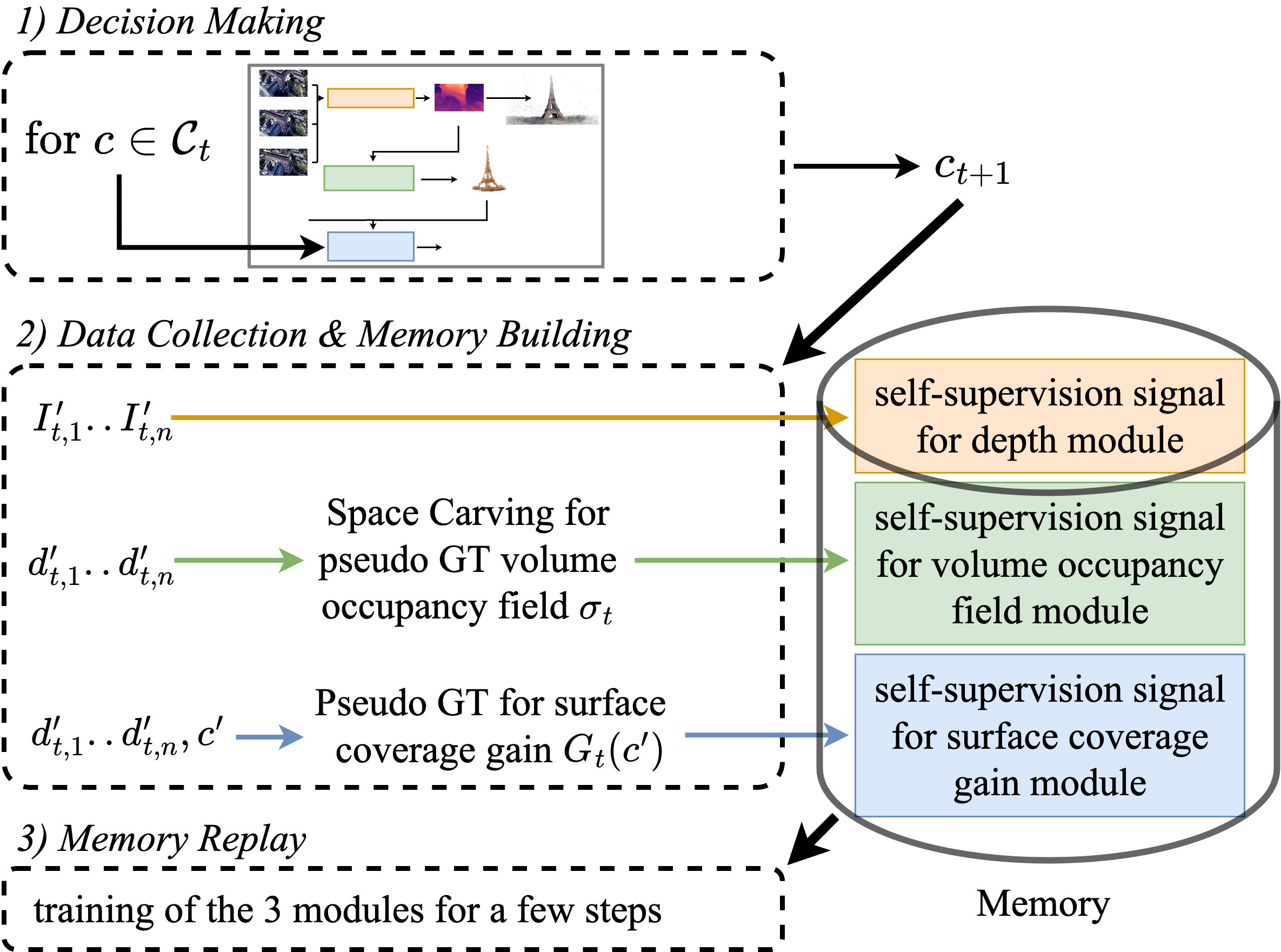}
  \caption{\textbf{Our architecture and the three steps of our self-supervised procedure.}}
  \label{fig:pipeline}
\end{figure*}

The general aim of Next Best View is to identify the next most informative sensor position(s) for reconstructing a 3D object or scene efficiently and accurately. Like previous works~\cite{guedon-nips22-scone-surface-coverage}, we look for the view that increases the most the total coverage of the scene surface. Optimizing such criterion makes sure we do not miss parts of the target scene.

We denote the set of occupied points in the scene by $\chi \subset \IR^3$; its boundary $\partial\chi$ is made of the surface points of the scene. During the exploration, at any time step $t\geq 0$, our method has built a partial knowledge of the scene: It has captured \emph{observations} $(I_0, ..., I_t)$ from the \emph{poses} $(c_0, ..., c_t)$ it visited. The 6D poses $c_i=(c_i^\pos, c_i^\rot)\in\calC:=\IR^3\times SO(3)$ encode both the position and the orientation of the sensor. In our case, observations $I_i$ are RGB images. 

To solve the NBV problem, we want to build a model that takes as inputs $(c_0, ..., c_t)$ and $(I_0,...,I_t)$ and predicts the next sensor pose $c_{t+1}$ that will maximize the number of new visible surface points, \ie, points in $\partial\chi$ that will be visible in the observation $I_{t+1}$ but not in the previous observations $I_0, ..., I_t$. We call the number of new visible surface points the \emph{surface coverage gain}. We assume the method is provided a 3D bounding box to delimit the part of the scene it should reconstruct.

\section{Method}
\label{sec:method}

Figure~\ref{fig:pipeline} gives an overview of our pipeline and our self-supervised online learning procedure. During online exploration, we perform a training iteration at each time step $t$ which consists in three steps. 

First, during the \emph{Decision Making step}, 
we select the next best camera pose to explore the scene by running our three modules:  the \emph{depth module} predicts the depth map for the current frame from the last capture frames, which is added to a point cloud that represents the scene. This point cloud is used by the \emph{volume occupancy module} to predict a volume occupancy field, which is in turn used by the \emph{surface coverage gain module} to compute the surface coverage gain of a given camera pose. We use this last module to find a camera pose around the current pose that optimizes this surface coverage gain. 

Second, the \emph{Data Collection \& Memory Building} step, during which the camera moves toward the camera pose previously predicted, creates a self-supervision signal for all three modules and stores these signals into the Memory. 

Third and last, the \emph{Memory Replay} step selects randomly supervision data stored into the Memory and updates the weights of each of the three modules.

We detail below our architecture and the three steps of our training procedure.

\subsection{Architecture}\label{sec:archi}

\paragraph{Depth module.} The goal of this module is to reconstruct the surface points observed by the RGB camera in real time during the exploration. To this end, it takes as input a sequence of recently captured images $I_t, I_{t-1}, .., I_{t-m}$ as well as the corresponding camera poses $c_t, c_{t-1}, .., c_{t-m}$ with $0 \leq m\leq t$  and predicts the depth map $d_t$ corresponding to the latest observation $I_t$. 

We follow Watson~\etal~\cite{watson-cvpr21-temporal-opportunist-manydepth} and build this module around a cost-volume feature. We first use pretrained ResNet18~\cite{he-cvpr16-deepresiduallearning} layers to extract features $f_i$ from images $I_i$. We define a set of ordered planes perpendicular to the optical axis at $I_t$ with depths linearly spaced between extremal values.
Then, for each depth plane, we use the camera poses to warp the features $f_{t-j}, 0<j\leq m$ to the image coordinate system of the camera $c_t$, and compute the pixelwise L1-norm between the warped features and $f_t$. This results in a cost volume that encodes for every pixel the likelihood of each depth plane to be the correct depth. 
We implement this depth prediction with a U-Net architecture~\cite{ronneberger-15-unet} similar to~\cite{watson-cvpr21-temporal-opportunist-manydepth} that takes as inputs $f_t$ and the cost volume to recover $d_t$. More details can be found in~\cite{watson-cvpr21-temporal-opportunist-manydepth}.
Contrary to~\cite{watson-cvpr21-temporal-opportunist-manydepth}, we suppose the camera poses to be known for faster convergence. We use $m=2$ in our experiments. In practice, the most recent images of our online learning correspond to images captured along the way between two predicted poses $c_t$ and $c_{t+1}$ so we use them instead of $I_0, .., I_t$.

We then backproject the depth map $d_t$ in 3D, filter the point cloud and concatenate it to the previous points obtained from $d_0, .., d_{t-1}$. We filter points associated to strong gradients in the depth map, which we observed are likely to yield wrong 3D points: We remove points based on their value for the edge-aware smoothness loss appearing in~\cite{watson-cvpr21-temporal-opportunist-manydepth,godard-cvpr17-unsupervisedmonoculardepth,heise-iccv13-pmhuber} that we also use for training. We hypothesize such outliers are linked to the module incapacity to output sudden changes in depth, thus resulting in over-smooth depth maps. We denote by $S_t$ the reconstructed surface point cloud resulting from all previous projections.

\begin{figure}
  \centering

  \begin{subfigure}{0.23\linewidth}
    \includegraphics[width=\linewidth]{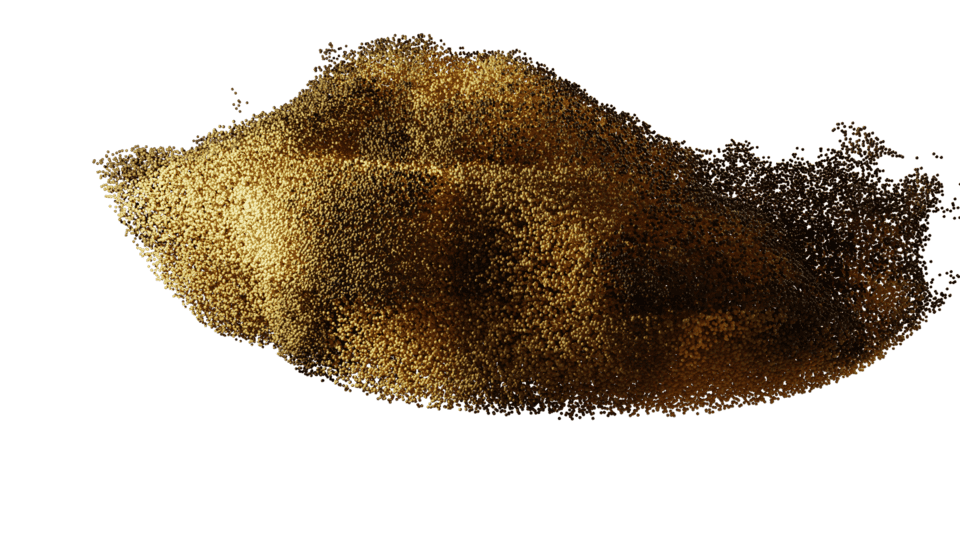}
    \label{fig:occ-b}
  \end{subfigure}
  \hfill
  \begin{subfigure}{0.23\linewidth}
    \includegraphics[width=\linewidth]{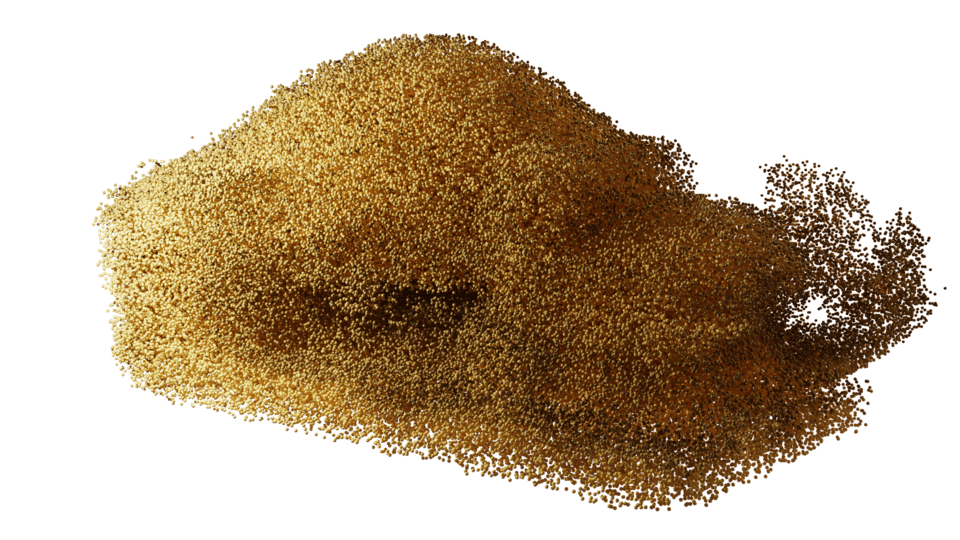}
    \label{fig:occ-c}
  \end{subfigure}
  \hfill
  \begin{subfigure}{0.23\linewidth}
    \includegraphics[width=\linewidth]{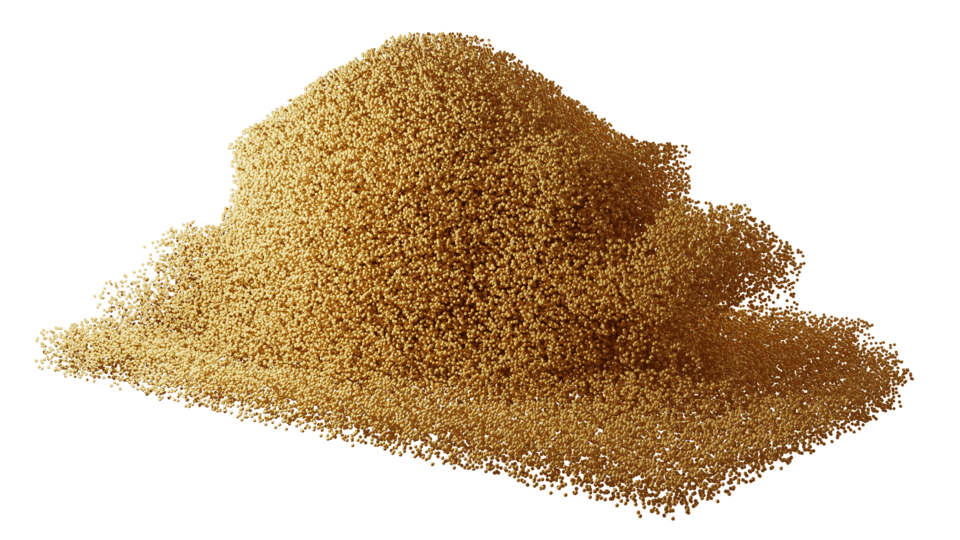}
    \label{fig:occ-d}
  \end{subfigure}
  \hfill
    \begin{subfigure}{0.23\linewidth}
    \includegraphics[width=\linewidth]{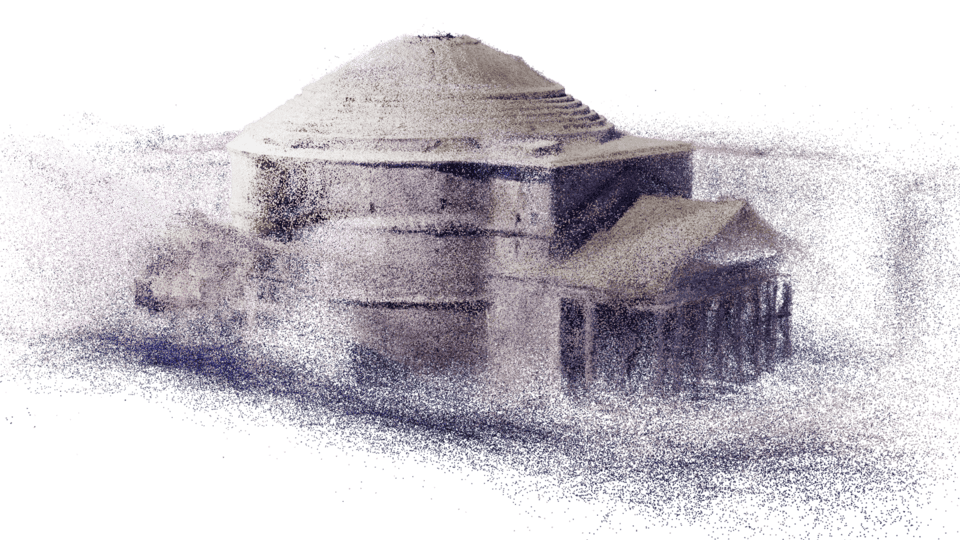}
    \label{fig:occ-a}
  \end{subfigure}\\
  \begin{subfigure}{0.23\linewidth}
    \includegraphics[width=\linewidth]{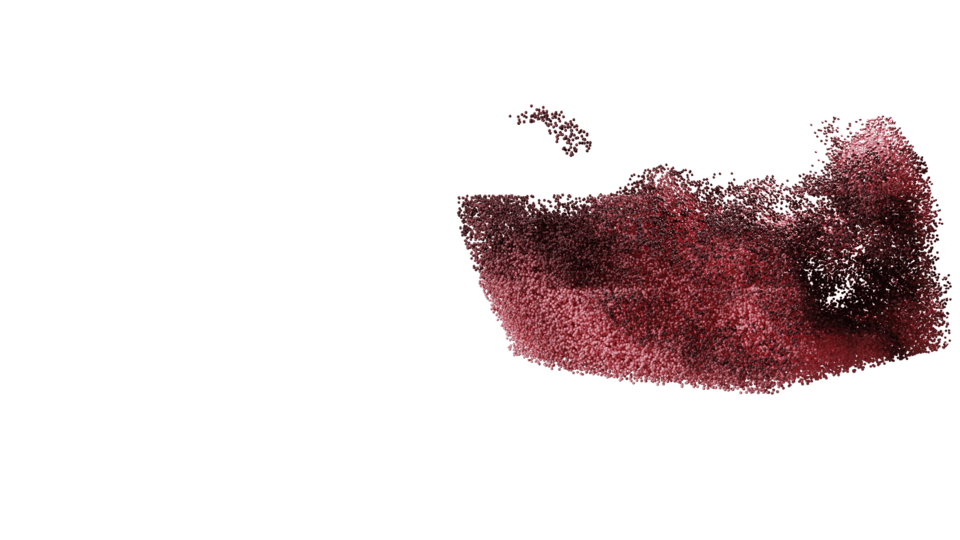}
    \caption{\scriptsize Volume occupancy at $t_0$}
    \label{fig:occ-b-bis}
  \end{subfigure}
    \hfill
  \begin{subfigure}{0.23\linewidth}
    \includegraphics[width=\linewidth]{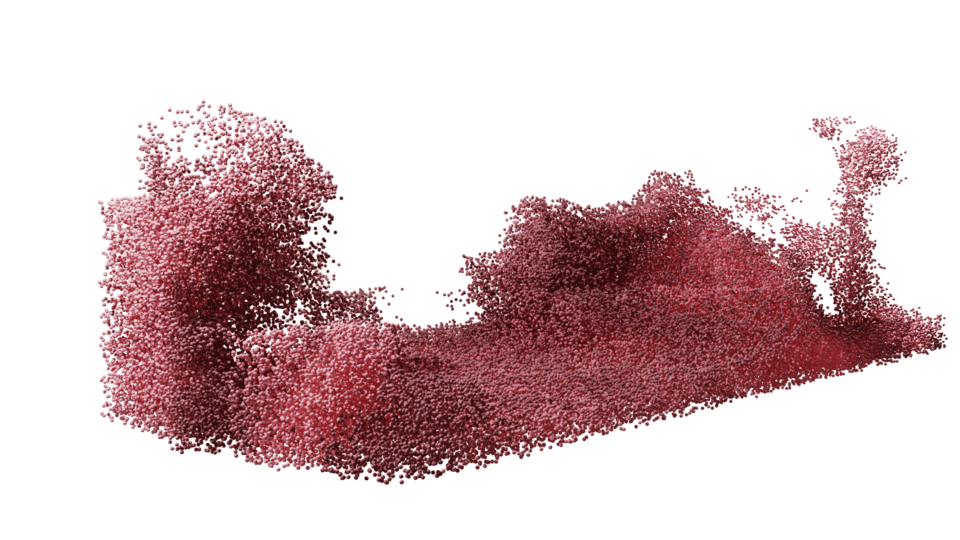}
    \caption{\scriptsize Volume occupancy at $t_1 > t_0$}
    \label{fig:occ-c-bis}
  \end{subfigure}
  \hfill
  \begin{subfigure}{0.23\linewidth}
    \includegraphics[width=\linewidth]{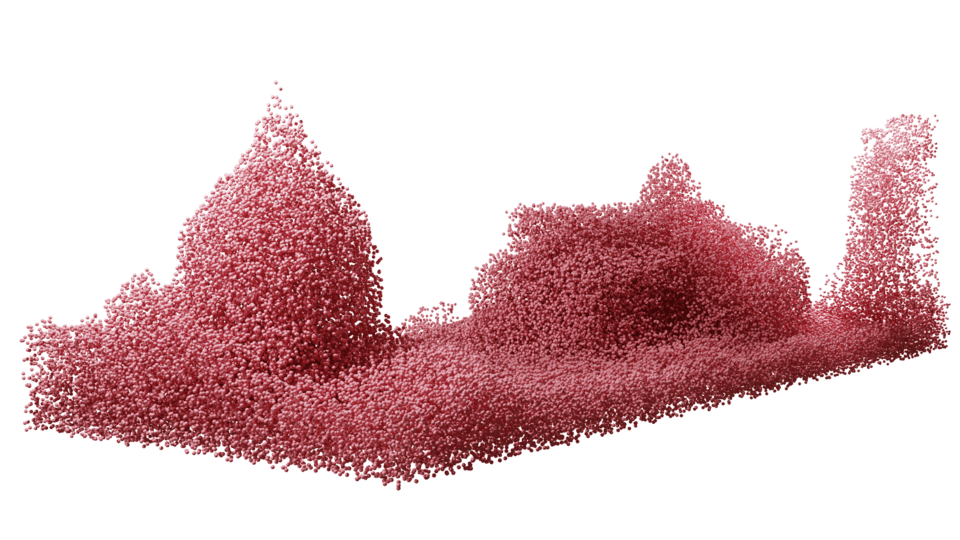}
    \caption{\scriptsize Volume occupancy at $t_2 > t_1$}
    \label{fig:occ-d-bis}
  \end{subfigure}
  \hfill
  \begin{subfigure}{0.23\linewidth}
    \includegraphics[width=\linewidth]{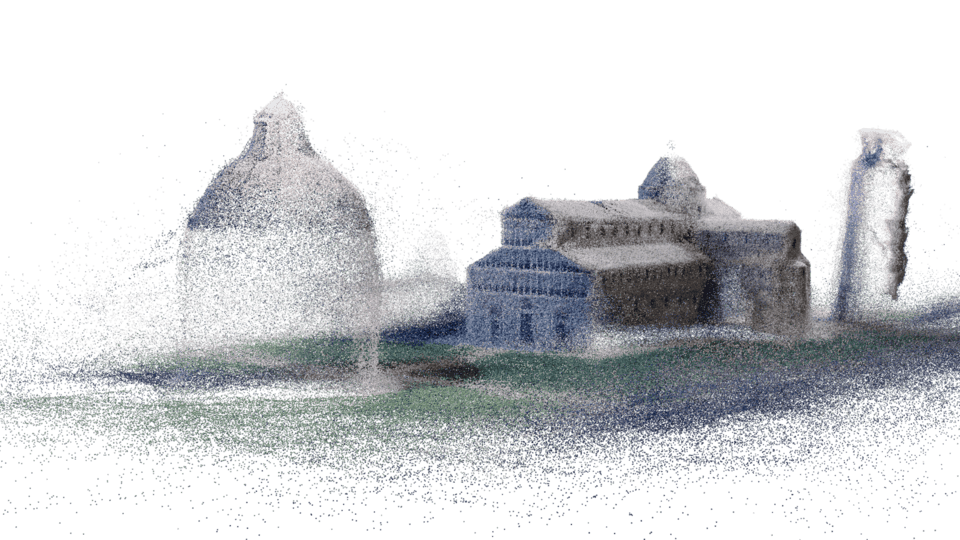}
    \caption{\scriptsize Reconstructed surface}
    \label{fig:occ-a-bis}
  \end{subfigure}
  \caption{\textbf{Evolution of the volume occupancy field and final surface} estimated by MACARONS on two examples.}
  \label{fig:volume-occupancy-probability}
\end{figure}

\paragraph{Volume occupancy module.} This module computes a ``volume occupancy field'' $\sigma_t$ from the predicted depth maps.
Given a 3D point $p$, $\sigma_t(p)=0$ indicates that the module is confident the point is empty; $\sigma_t(p)=1$ indicates that the module is confident the point is occupied. As shown in Figure~\ref{fig:volume-occupancy-probability}, during exploration, $\sigma_t(p)$ evolves as the module becomes more and more confident that $p$ is empty or occupied.

We implement this module using a Transformer~\cite{vaswani-nips17-attentionisallyouneed} taking as input the point $p$, the surface point cloud $S_t$ and previous poses $c_i$, and outputting a scalar value in $[0, 1]$. The exact architecture is provided in the appendix. This volumetric representation is convenient to build a NBV prediction model that scales to large environments. Indeed, it has a virtually infinite resolution and can handle arbitrarily large point clouds without failing to encode fine details since it uses mostly local features at different scales to compute the probability of a 3D point to be occupied. 

\paragraph{Surface coverage gain module.}
The final module computes the surface coverage gain of a given camera pose $c$ based on the predicted occupancy field, as proposed by~\cite{guedon-nips22-scone-surface-coverage} but with key modifications.

Similar to~\cite{guedon-nips22-scone-surface-coverage}, given a time step $t$, a camera pose $c$ and a 3D point $p$, we define the visibility gain $g_t(c;p)$ as a scalar function in $[0, 1]$ such that values close to 1 correspond to occupied points that will become visible through $c$ and values close to 0 correspond to points not newly visible through $c$. In particular, the latter includes points with low occupancy, points not visible from $c$ or points already visible from prior poses. We model this function using a Transformer-based architecture accounting for both the predicted occupancy and the camera history. 

Specifically, we first sample $N$ random points $(p_j)_{1\leq j\leq N}$ in the field of view $F_c$ of camera $c$ with probabilities proportional to $\sigma_t(p_j)$ using inverse transform sampling. Second, we represent the camera history of previous camera poses $c_0, ..., c_t$ for each point $p_j$ by projecting them on the sphere centered on $p_j$ and encoding the result into a spherical harmonic feature we denote by $h_t(p_j)$. Finally, we feed the camera pose $c$, a 3D point $p_j$, its occupancy $\sigma_t(p_j)$ as well as the camera history feature $h_t(p_j)$ to the Transformer predicting the corresponding visibility gain $g_t(c;p_j)$. Note that the self-attention mechanism is useful to deal with potential occlusions between points. 

The visibility gains of all points are aggregated using a Monte Carlo integration to estimate the surface coverage gain $G_t(c)$ of any camera pose $c$:%
\begin{equation}
    G_t(c) = V_c \cdot \frac{1}{N} \sum_{j=1}^N g_t(c; p_j) \cdot l(c; p_j) \> ,
    \label{coverage_gain}
\end{equation}
where $V_c$ and $l(c; p_j)$ are two key quantities we introduce compared to the original formula of~\cite{guedon-nips22-scone-surface-coverage}. First, we multiply the sum by $V_c = \int_{F_c} \sigma_t(p) dp$ (\ie, the volume of occupied points seen from $c$) to account for its variability between different camera poses, which is typically strong for complex 3D scenes. Second, since the density of surface points visible in images decrease with the distance between the surface and the camera, we also weight the visibility gains with factor $l(c; p_j) = \min(1/\|c^\pos-p_j\|^2,\tau)$ inversely proportional to the distance between the camera center and the point, to give less importance to points far away from the camera. We also made several minor improvements to the computation of the surface coverage gain, which we detail in the appendix.

\subsection{Decision Making: Predicting the NBV}

At any time $t$, the Decision Making step simply consists in applying successively the three modules of the model, as described in~\Cref{sec:archi}.
Consequently, we first apply the depth prediction module on the current frame $I_t$ and use the resulting depth map $d_t$ to update the surface point cloud $S_t$. Then, for a set of candidate camera poses $\calC_t \subset \calC$, we apply the other modules to compute in real time the volume occupancy field and estimate the surface coverage gain $G_t(c)$ of all camera poses $c\in\calC_t$.
In practice, we build $\calC_t$ by simply sampling around the current camera pose but more complex strategies could be developed.
We select the NBV as the camera pose with the highest coverage gain:
\begin{equation}
    c_{t+1} = \argmax_{c \in \calC_t} G_t(c) \> .
\end{equation}

We do not compute gradients nor update the weights of the model during the Decision Making step. Indeed, since the camera visits only one of the candidate camera poses at next time step $t+1$, we do not gather data about all neighbors. Consequently, we are not able to build a self-supervision signal involving every neighbor. As we explain in the next subsection, we build a self-supervision signal to learn coverage gain from RGB images only by exploiting the camera trajectory between poses $c_t$ and $c_{t+1}$.

\subsection{Data Collection \& Memory Building}
\begin{figure*}
  \centering
  \begin{subfigure}{0.23\linewidth}
    \includegraphics[width=\linewidth]{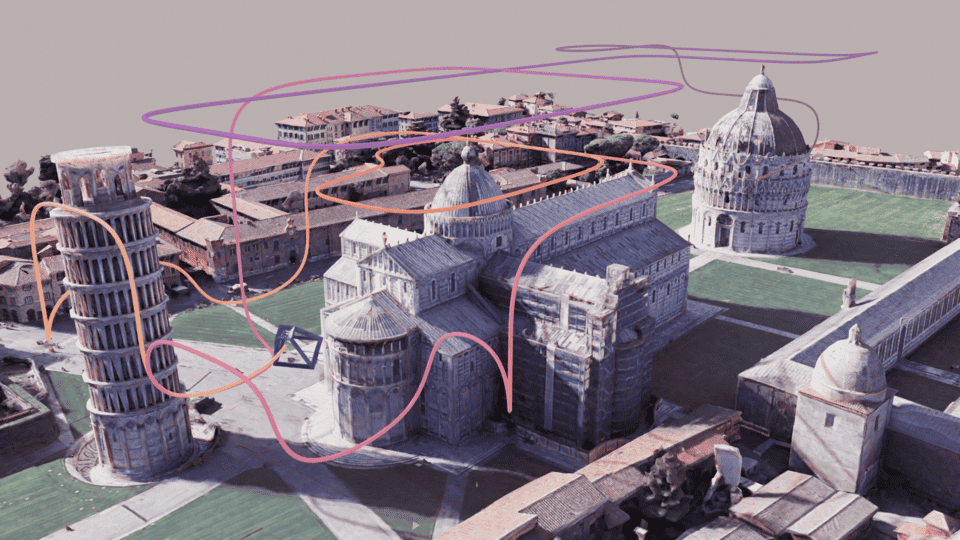}
    \caption{Pisa Cathedral}
  \end{subfigure}
  \hfill
  \begin{subfigure}{0.23\linewidth}
    \includegraphics[width=\linewidth]{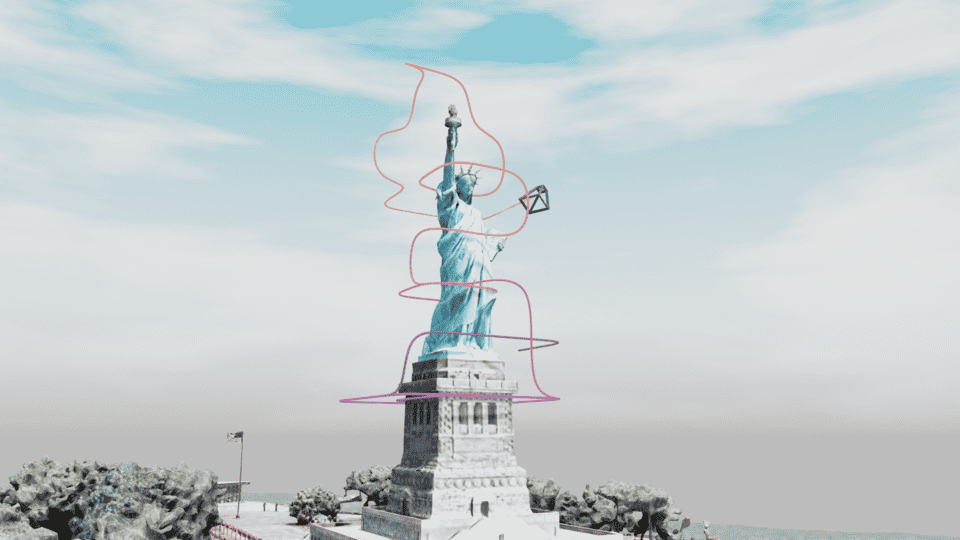}
    \caption{Statue of Liberty}
  \end{subfigure}
  \hfill
  \begin{subfigure}{0.23\linewidth}
    \includegraphics[width=\linewidth]{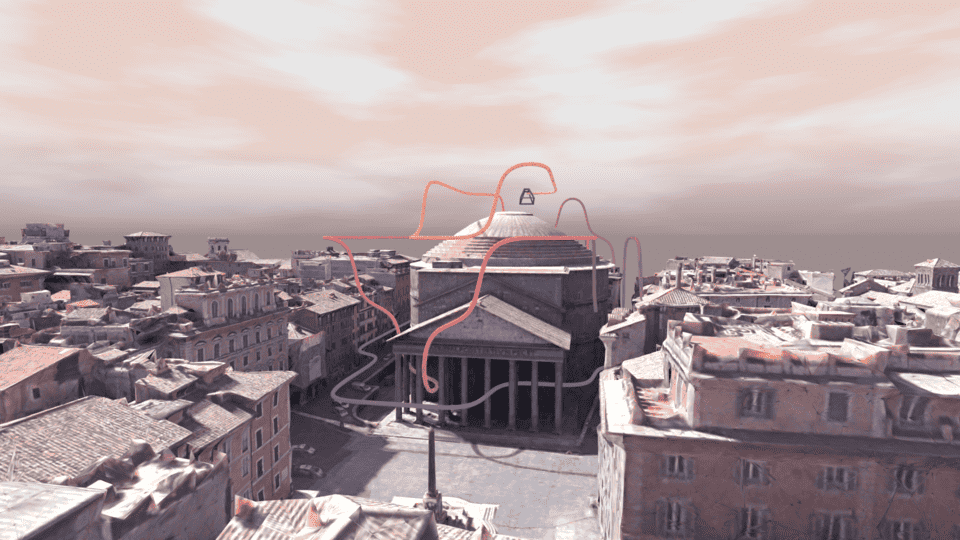}
    \caption{Pantheon}
  \end{subfigure}
  \hfill
  \begin{subfigure}{0.23\linewidth}
    \includegraphics[width=\linewidth]{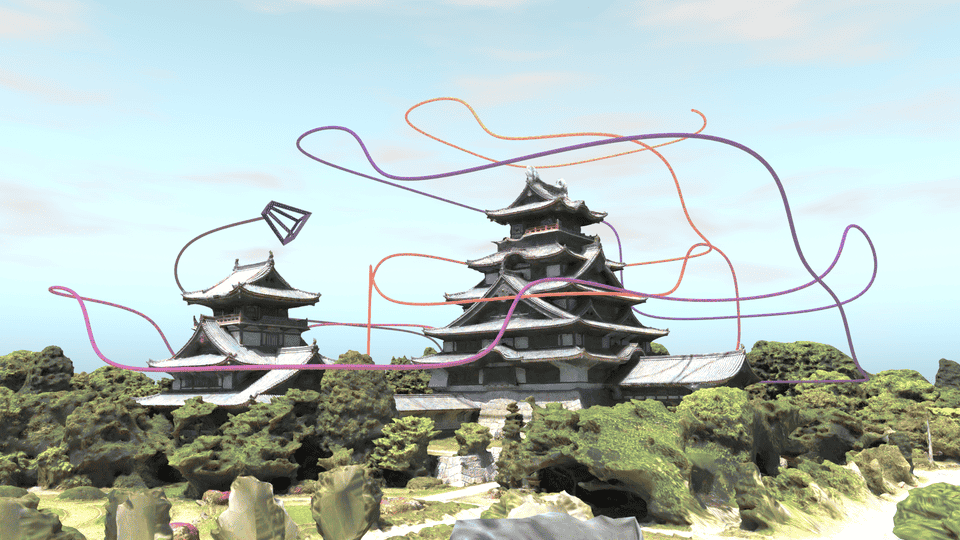}
    \caption{Fushimi Castle}
  \end{subfigure} \\
  \begin{subfigure}{0.23\linewidth}
    \includegraphics[width=\linewidth]{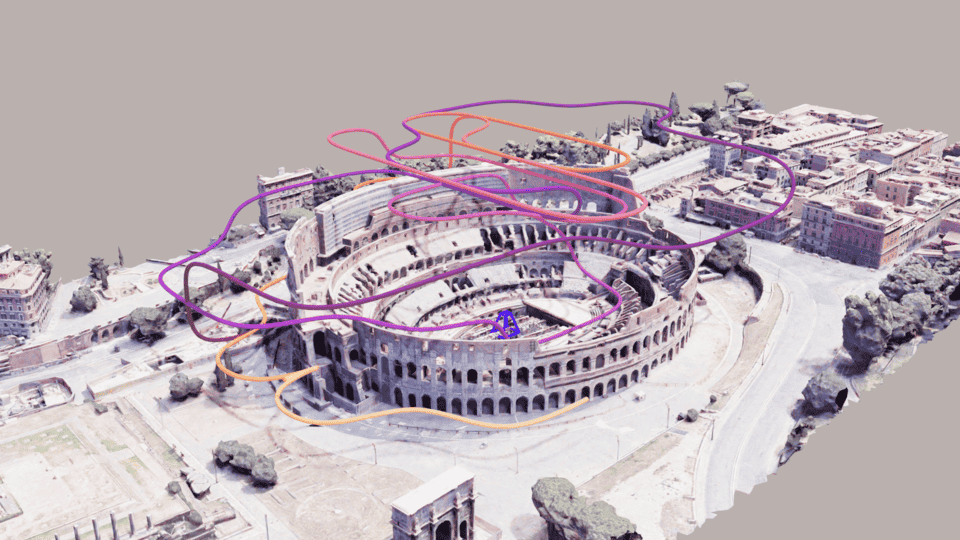}
    \caption{Colosseum}
  \end{subfigure}
  \hfill
  \begin{subfigure}{0.23\linewidth}
    \includegraphics[width=\linewidth]{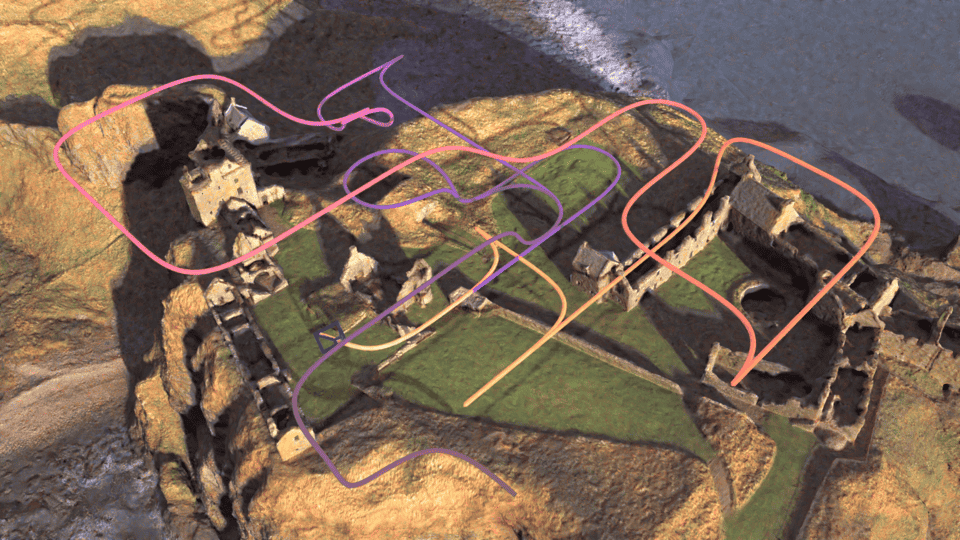}
    \caption{Dunnottar Castle}
  \end{subfigure}
  \hfill
  \begin{subfigure}{0.23\linewidth}
    \includegraphics[width=\linewidth]{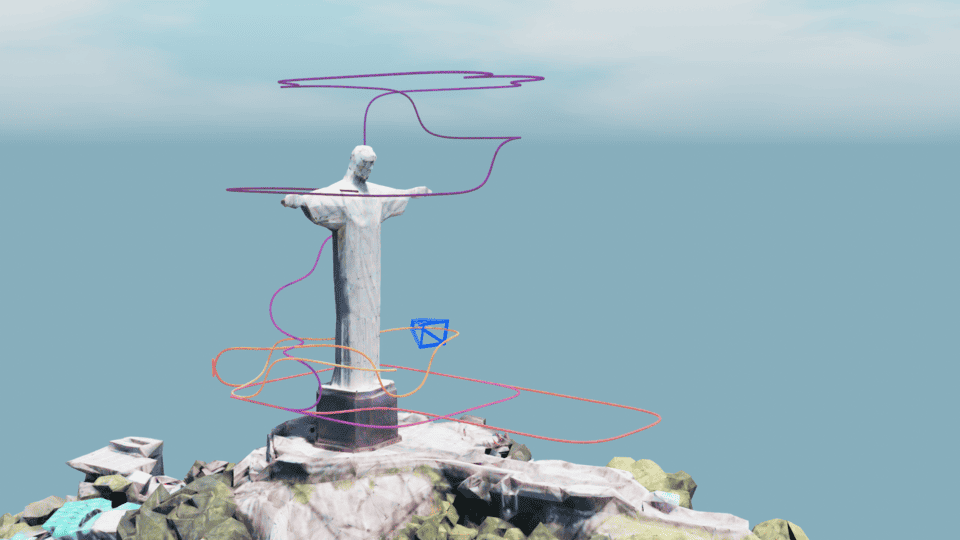}
    \caption{Christ the Redeemer}
  \end{subfigure}
  \hfill
  \begin{subfigure}{0.23\linewidth}
    \includegraphics[width=\linewidth]{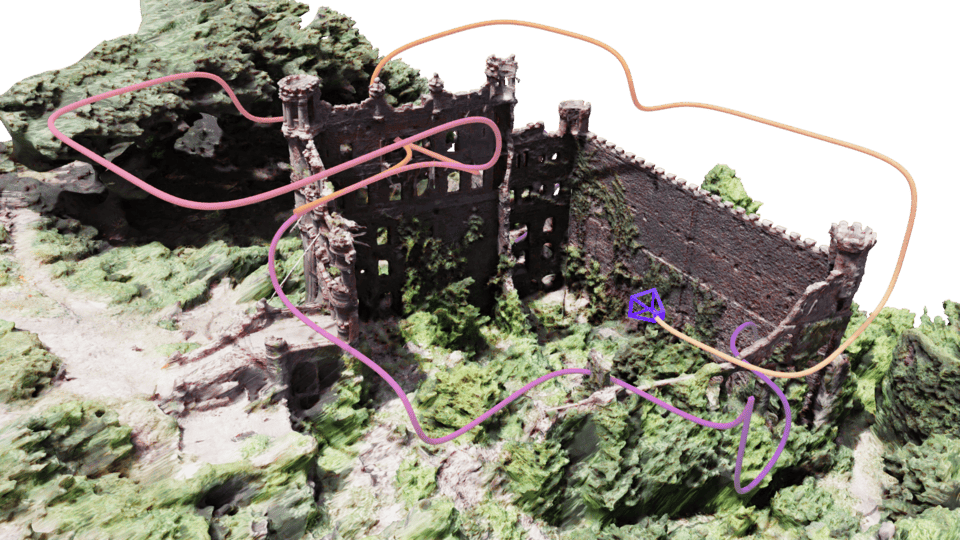}
    \caption{Bannerman Castle}
  \end{subfigure}
  \begin{subfigure}{0.23\linewidth}
    \includegraphics[width=\linewidth]{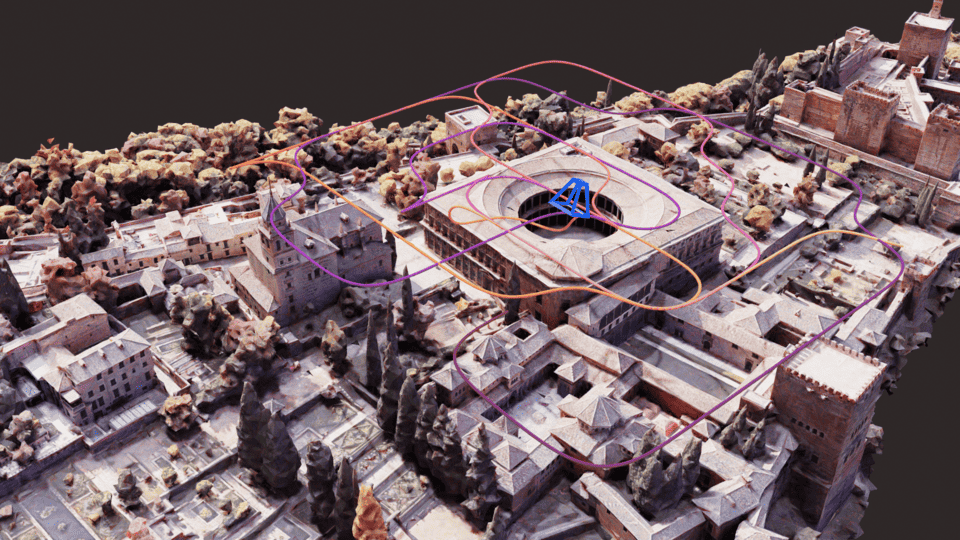}
    \caption{Alhambra Palace}
  \end{subfigure}
  \hfill
  \begin{subfigure}{0.23\linewidth}
    \includegraphics[width=\linewidth]{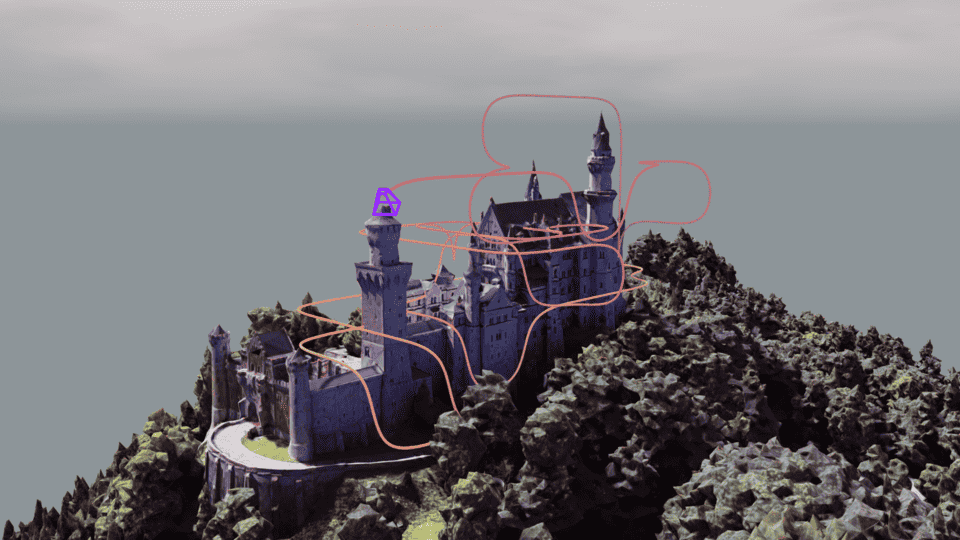}
    \caption{Neuschwanstein Castle}
  \end{subfigure}
  \hfill
  \begin{subfigure}{0.23\linewidth}
    \includegraphics[width=\linewidth]{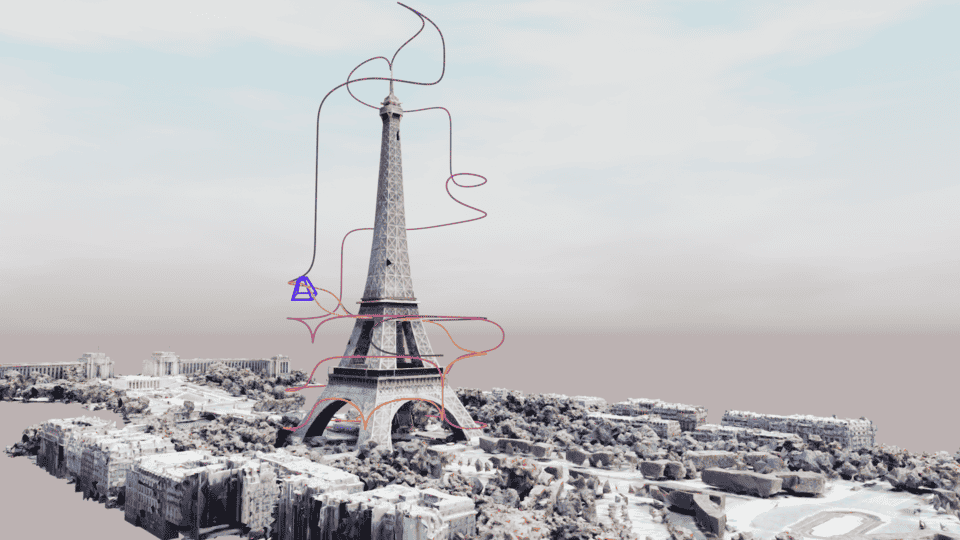}
    \caption{Eiffel Tower}
  \end{subfigure}
  \hfill
  \begin{subfigure}{0.23\linewidth}
    \includegraphics[width=\linewidth]{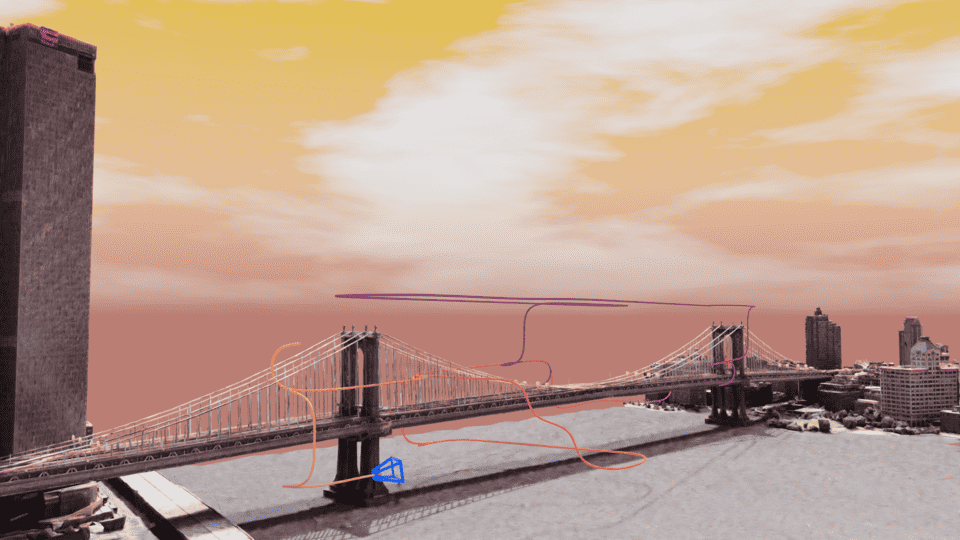}
    \caption{Manhattan Bridge}
  \end{subfigure}
  \caption{{\bf Trajectories computed in real-time by MACARONS in large 3D structures.} At each time step $t$, MACARONS predicts a NBV and builds trajectories that consistently cover most of the surface of the scene. MACARONS performed 100 NBV iterations in these images.}
  \label{fig:supp-mat-trajectory}
\end{figure*}  

During Data Collection \& Memory Building, we move the camera to the next pose $c_{t+1}$. This is done by simple linear interpolation between $c_t$ and $c_{t+1}$ and capture $n$ images along the way, including the image $I_{t+1}$ captured from the camera pose $c_{t+1}$. 
We denote these images by $I'_{t,1}, ..., I'_{t,n}$ so  $I'_{t, n} = I_{t+1}$, and write $I'_{t, 0} := I_t$.

Then, we collect a self-supervision signal for each of the three modules, which we store in the Memory. Some of the previous signals can be discarded at the same time, depending on the size of the Memory.

\paragraph{Depth module.} We simply store the consecutive frames $I'_{t,1}, ..., I'_{t,n}$, which we will use to train the module in a standard self-supervised fashion.

\paragraph{Volume occupancy module.} We rely on Space Carving~\cite{kutulakos00space} using the predicted depth maps to create a supervision signal to train the prediction of the volume occupancy field. Our key idea is as follows: When the whole surface of the scene is covered with depth maps, a 3D point $p \in \IR^3$ is occupied iff for any depth map $d$ containing $p$ in its field of view, $p$ is located behind the surface visible in $d$. Consequently, if we had images covering the entire scene and their corresponding depth maps, we could  compute the complete occupancy field of the scene by removing all points that are not located behind depth maps. 

In practice, we only have access to the depth maps $d_{t, 1}', ..., d_{t,n}'$ predicted for the images captured so far. We can still compute an intermediate occupancy field, which is an approximation but can be used as supervision signal. Since it is not reliable far away from the depth maps when the whole surface has not been covered, we only sample points around the newly reconstructed surface within a margin that increases with the total number of depth maps. 

Finally, we store in the Memory some of the sampled points with their occupancy value for future supervision, and update the value of points already stored in the Memory.

\paragraph{Surface coverage gain module.} The process to build supervision values for training the surface coverage gain prediction is as follows. Using the data collected at time steps $i\leq t$, we apply the surface coverage gain module to predict the surface coverage gain of camera poses $c'_{t,1}, ..., c'_{t,n-1}$. Then, for each $c'_{t,i}$, we compute a supervision value for the coverage gain by counting the number of new visible surface points appearing in the depth map $d'_{t, i}$.  
We consider a surface point to be new if its distance to the previous reconstructed point cloud $S_t$ is at least $\epsilon$, where $\epsilon$ is a hyperparameter used for coverage evaluation.

We finally update the surface point cloud $S_t$ stored in the Memory. We also store the poses $c'_{t,i}$ and the depth maps $d'_{t,i}$, in order to recompute supervision values for surface coverage gain when sampling from the Memory.

\subsection{Memory Replay}

During this step, we randomly sample the data stored in the Memory to train each of the modules as follows. We add to these samples the newly acquired data, to make sure the model learns from the current state of the scene. The more memory replay iterations, the faster the model learns and converges but the slower it explores.

\paragraph{Depth module.} We follow a standard loss from self-supervised monocular depth prediction literature~\cite{watson-cvpr21-temporal-opportunist-manydepth,godard-iccv19-digginginto,godard-cvpr17-unsupervisedmonoculardepth} to train the depth prediction module from RGB images. The only difference is that in our case, we use multiple input images. We thus predict the depth map for the current image from the $m$ previously captured images. The loss compares these images after warping using the predicted depth map with the same reconstruction loss as~\cite{watson-cvpr21-temporal-opportunist-manydepth,godard-iccv19-digginginto,godard-cvpr17-unsupervisedmonoculardepth}, which is a combination of SSIM, L1 and an edge-aware smoothness loss. We also include in this loss the next image as suggested in~\cite{watson-cvpr21-temporal-opportunist-manydepth}, since it greatly improves performance.

\paragraph{Volume occupancy module.}

We train this module by comparing its predictions, computed from $S_t$ and the previous camera poses, to the updated carved occupancy field computed from the newly acquired data with the MSE loss. 
\begin{figure*}
  \centering
  \begin{subfigure}{0.23\linewidth} 
    \includegraphics[width=\linewidth]{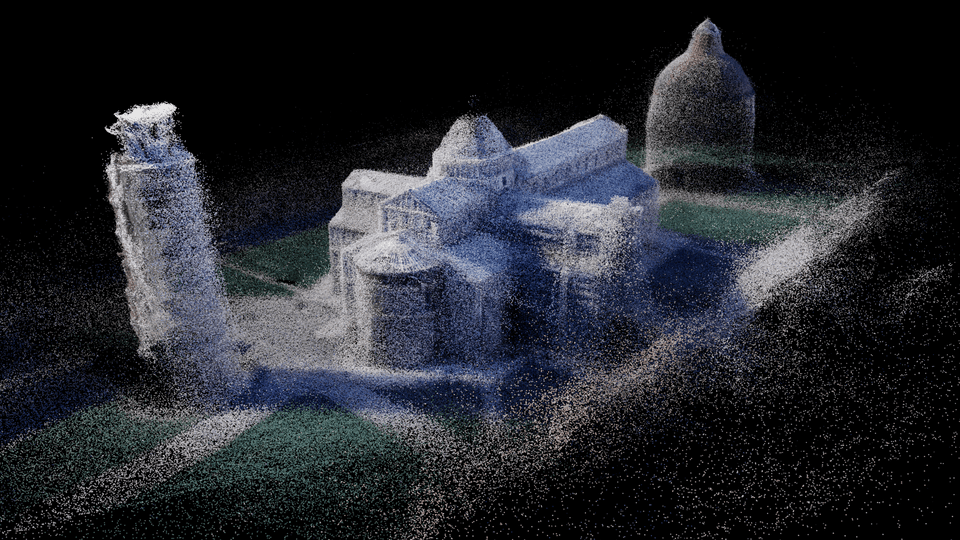}
    \caption{\scriptsize Pisa Cathedral}
    \label{fig:surface-a}
  \end{subfigure}
  \hfill
  \begin{subfigure}{0.23\linewidth}
    \includegraphics[width=\linewidth]{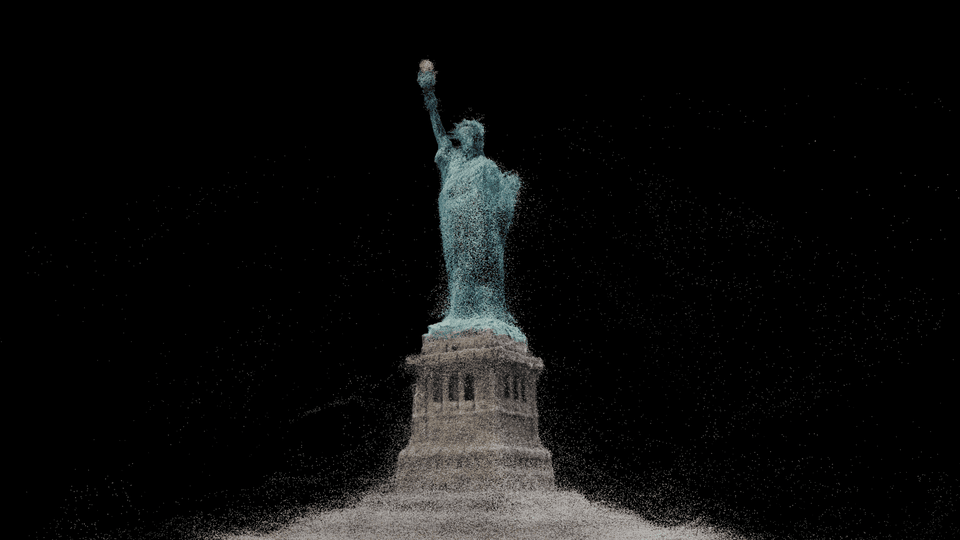}
    \caption{\scriptsize Statue of Liberty}
    \label{fig:surface-b}
  \end{subfigure}
  \hfill
  \begin{subfigure}{0.23\linewidth}
    \includegraphics[width=\linewidth]{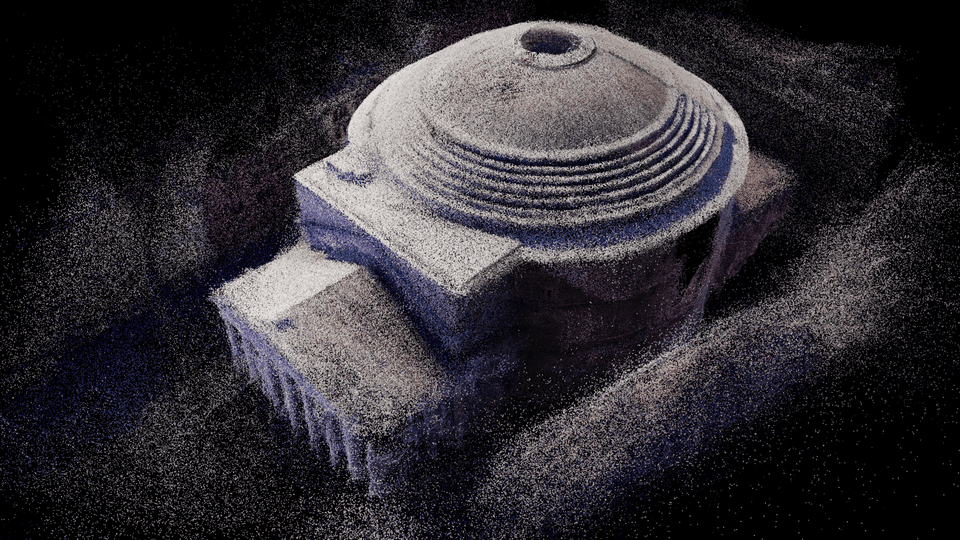}
    \caption{\scriptsize Pantheon}
    \label{fig:surface-c}
  \end{subfigure}
  \hfill
  \begin{subfigure}{0.23\linewidth}
    \includegraphics[width=\linewidth]{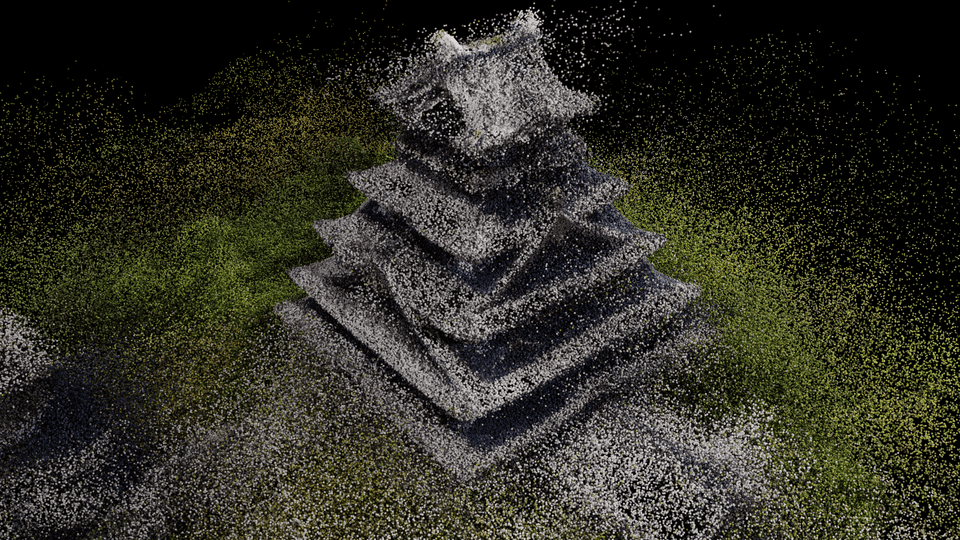}
    \caption{\scriptsize Fushimi Castle}
    \label{fig:surface-d}
  \end{subfigure} 
  \hfill
  \begin{subfigure}{0.23\linewidth}
    \includegraphics[width=\linewidth]{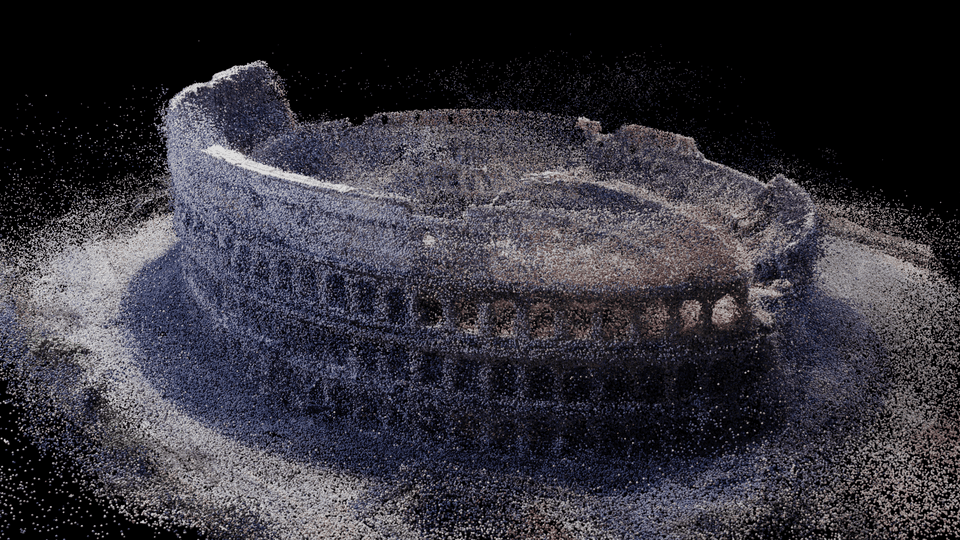}
    \caption{\scriptsize Colosseum}
    \label{fig:surface-e}
  \end{subfigure}
  \hfill
  \begin{subfigure}{0.23\linewidth}
    \includegraphics[width=\linewidth]{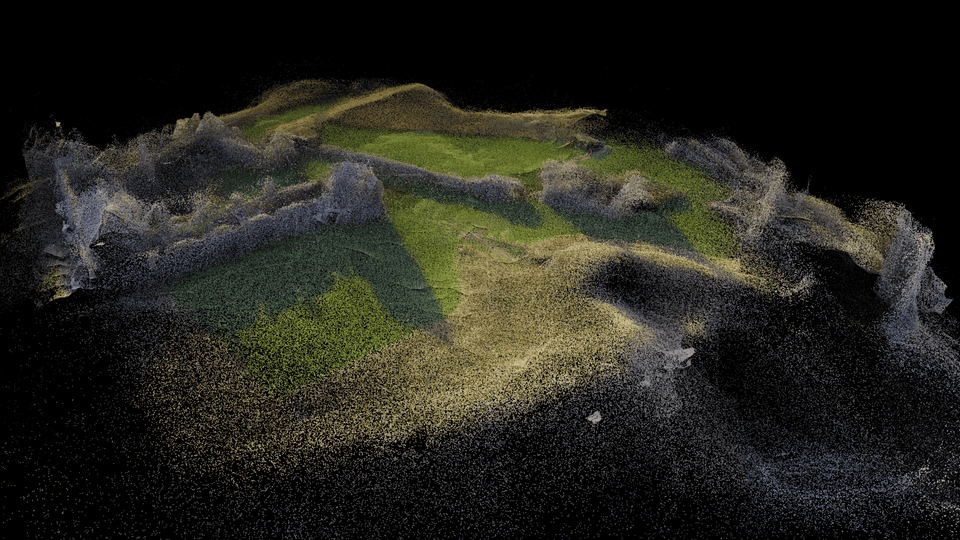}
    \caption{\scriptsize Dunnottar Castle}
    \label{fig:surface-f}
  \end{subfigure}
  \hfill
  \begin{subfigure}{0.23\linewidth}
    \includegraphics[width=\linewidth]{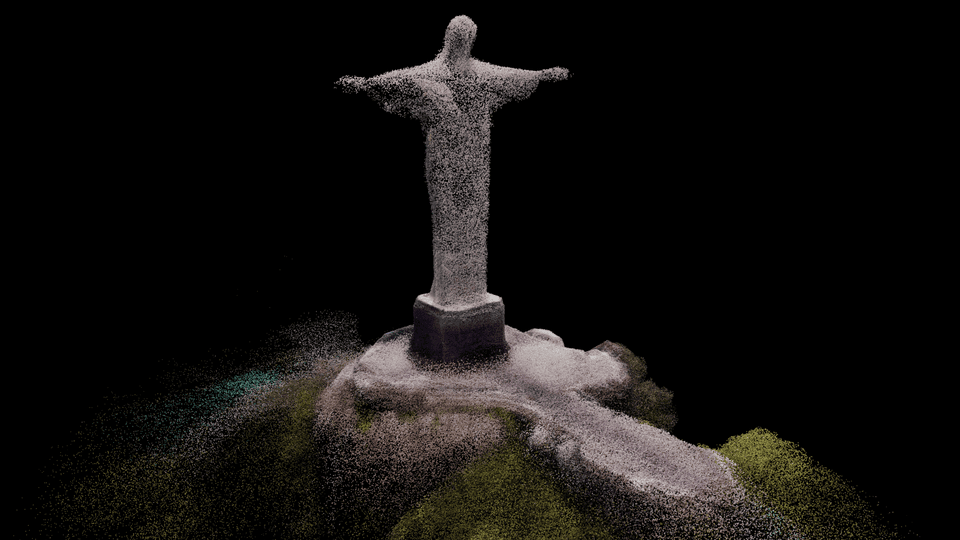}
    \caption{\scriptsize Christ the Redeemer}
    \label{fig:surface-g}
  \end{subfigure}
  \hfill
  \begin{subfigure}{0.23\linewidth}
    \includegraphics[width=\linewidth]{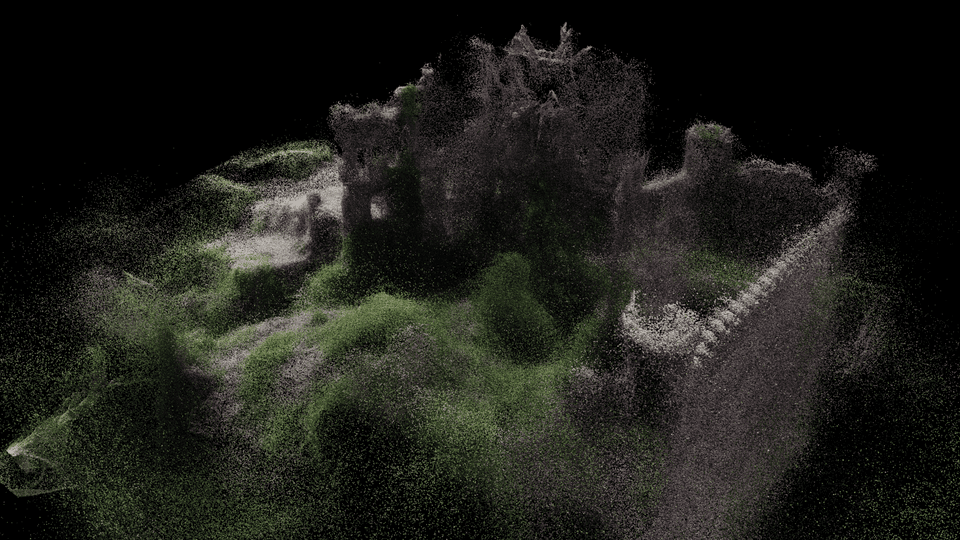}
    \caption{\scriptsize Bannerman Castle}
    \label{fig:surface-h}
  \end{subfigure}
  \hfill
  \begin{subfigure}{0.23\linewidth}
    \includegraphics[width=\linewidth]{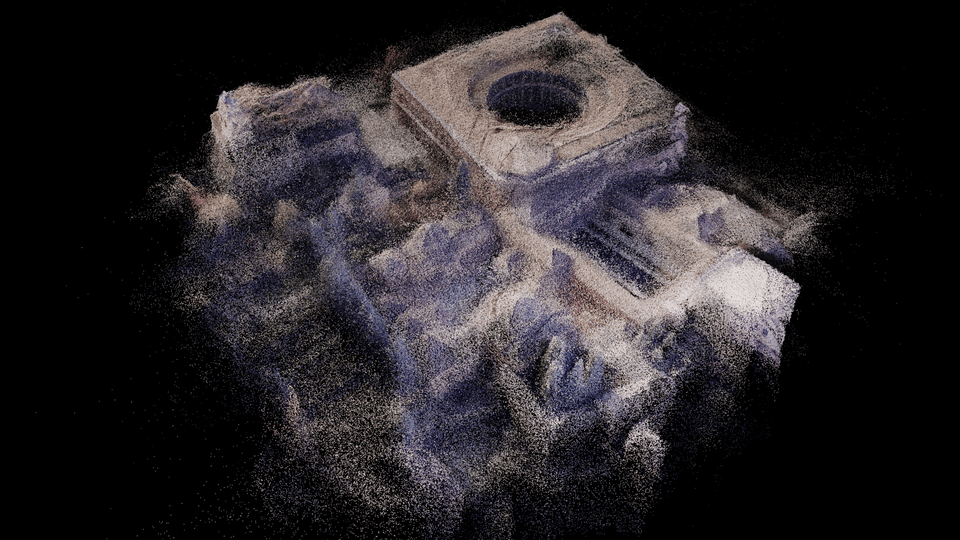}
    \caption{\scriptsize Alhambra Palace}
    \label{fig:surface-i}
  \end{subfigure}
  \hfill
  \begin{subfigure}{0.23\linewidth}
    \includegraphics[width=\linewidth]{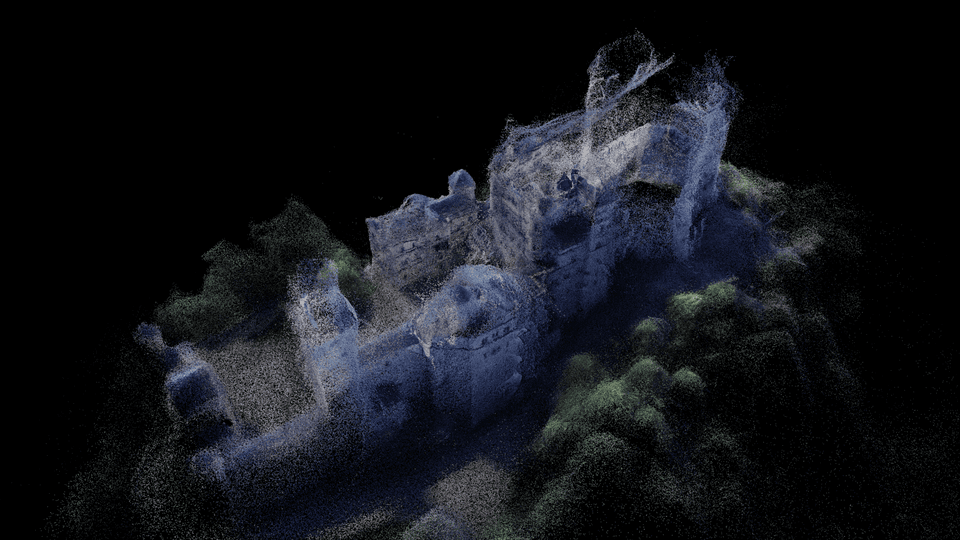}
    \caption{\scriptsize Neuschwanstein Castle}
    \label{fig:surface-j}
  \end{subfigure}
  \hfill
  \begin{subfigure}{0.23\linewidth}
    \includegraphics[width=\linewidth]{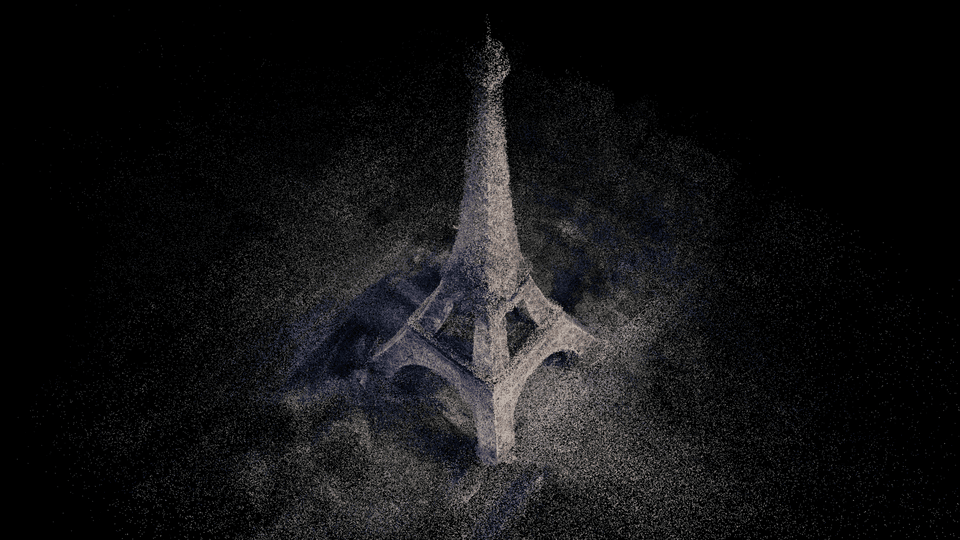}
    \caption{\scriptsize Eiffel Tower}
    \label{fig:surface-k}
  \end{subfigure}
  \hfill
  \begin{subfigure}{0.23\linewidth}
\includegraphics[width=\linewidth]{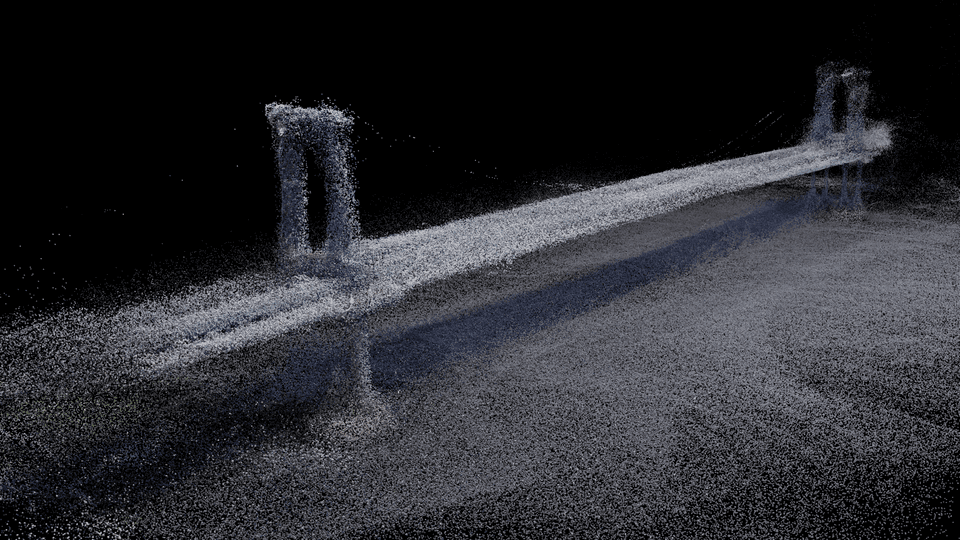}
    \caption{\scriptsize Manhattan Bridge}
    \label{fig:surface-l}
  \end{subfigure}
  \caption{{\bf Automated reconstruction of large 3D structures from RGB images with our approach MACARONS.} Our model reconstructs the surface in real time during exploration: We show here the reconstruction after 100 NBV iterations. The model has been trained on a set of previous scenes; all weights are frozen and we only perform inference computation.
  The first two rows depict scenes that were already seen by the model during its online, self-supervised training. The last row depicts scenes the model has never seen before. 
  MACARONS is not only able to reconstruct surfaces thanks to its depth prediction module (even for scenes it has never seen before), but is also able to optimize its path around the structure and consistently cover most of the surface of the scene thanks to its NBV prediction.}
  \label{fig:surface-reconstruction}
\end{figure*}  

\paragraph{Surface coverage gain module.}

We rely on a loss different from \cite{guedon-nips22-scone-surface-coverage} to train the surface coverage gain, which improves performance and interpretability.
\cite{guedon-nips22-scone-surface-coverage} showed that its formalism can estimate the surface coverage gain by integrating over the volume occupancy, but only to a scale factor that cannot be computed in closed form. Moreover, their training approach requires to have a dense set of cameras for each forward pass, since they compute the surface coverage gain as a distribution over the whole set of camera poses to compute their loss. They solve this requirement using many renderings of ShapeNet objects, but such a dense set is not available in our online self-supervised setting. Also, their normalization using softmax does not enforce the lowest visibility gain values to be close to zero.

Since the predicted coverage gains are supposed to be proportional to the real values, we propose a much simpler approach that consists in dividing both predicted and supervision coverage gains by their respective means on a potentially small set of cameras. We then compare these normalized coverage gains directly with a L1-norm. This simpler loss also enforces the lowest visibility gain values to be equal to zero. Overall, this loss function applies better constraints on the model to target meaningful visibility gains, and allows for training with less camera poses, which is essential to let our model learn in an online self-supervised fashion where only a few coverage gain values are available at real-time. Additional figures showing the proportionality of predicted and true coverage gains are available in the appendix.
\section{Experiments}
\label{sec:experiments}

\subsection{Implementation}


We implemented our model with PyTorch~\cite{paszke-nips19-pytorch} and use 3D data processing tools from PyTorch3D~\cite{ravi-arxiv20-accelerating3ddeeplearning}, such as ray-casting renderers to generate RGB images as inputs to our model.
MACARONS learns online to explore large, unknown environments thanks to its self-supervised pipeline that does not need any 3D input data; After being trained long enough, we can either freeze the weights and deactivate online learning to save computation time for future exploration, or let the model continue its training to further finetune it to novel scenes.
We perform online training with up to 4 GPUs Nvidia Tesla V100 SXM2 32 Go to let the model explore 4 different scenes in parallel and speed up the convergence, but we used a single GPU Nvidia GeForce GTX 1080 Ti for the inference experiments presented below.
In our experimental setup, after each NBV selection step, we perform 5 memory replay iterations for the depth module and up to 3 for the other modules. We provide extensive details in the appendix.

\subsection{Exploration of large 3D scenes}

%

\begin{table*}
  \centering
  {
  \scalebox{0.8}{
  \begin{tabular}{@{}lcccc@{}}
    \toprule
    3D scene & Random Walk & SCONE-Entropy~\cite{guedon-nips22-scone-surface-coverage} & SCONE~\cite{guedon-nips22-scone-surface-coverage} & MACARONS (Ours) \\
    \midrule
    Dunnottar Castle 
    & 0.330 $\pm$ 0.106 
    & 0.381 $\pm$ 0.041 
    & 0.650 $\pm$ 0.093 
    & \textbf{0.784} $\pm$ 0.025 \\
    Colosseum 
    & 0.312 $\pm$ 0.098
    & 0.477 $\pm$ 0.023
    & 0.532 $\pm$ 0.032
    & \textbf{0.629} $\pm$ 0.011 \\
    Bannerman Castle 
    & 0.316 $\pm$ 0.106
    & 0.534 $\pm$ 0.012
    & 0.552 $\pm$ 0.020
    & \textbf{0.716} $\pm$ 0.029 \\
    Pantheon 
    & 0.206 $\pm$ 0.064
    & 0.338 $\pm$ 0.030
    & 0.401 $\pm$ 0.030
    & \textbf{0.499} $\pm$ 0.019 \\
    Christ the Redeemer 
    & 0.518 $\pm$ 0.058
    & 0.836 $\pm$ 0.028
    & 0.833 $\pm$ 0.037
    & \textbf{0.865} $\pm$ 0.037\\
    Statue of Liberty 
    & 0.296 $\pm$ 0.188
    & 0.673 $\pm$ 0.025
    & 0.695 $\pm$ 0.020 
    & \textbf{0.708 }$\pm$ 0.014\\
    Pisa Cathedral 
    & 0.327 $\pm$ 0.121
    & 0.431 $\pm$ 0.024
    & 0.555 $\pm$ 0.033
    & \textbf{0.661} $\pm$ 0.008 \\
    Fushimi Castle 
    & 0.513 $\pm$ 0.091
    & 0.791 $\pm$ 0.020
    & 0.806 $\pm$ 0.029
    & \textbf{0.842} $\pm$ 0.019 \\
    \midrule
    Alhambra Palace 
    & 0.361 $\pm$ 0.045
    & 0.425 $\pm$ 0.046
    & 0.528 $\pm$ 0.030
    & \textbf{0.634} $\pm$ 0.019\\
    Neuschwanstein Castle 
    & 0.396 $\pm$ 0.090
    & 0.491 $\pm$ 0.039
    & 0.662 $\pm$ 0.041
    & \textbf{0.760} $\pm$ 0.020 \\
    Eiffel Tower 
    & 0.417 $\pm$ 0.074
    & 0.702 $\pm$ 0.016
    & 0.753 $\pm$ 0.013
    & \textbf{0.789} $\pm$ 0.010 \\
    Manhattan Bridge 
    & 0.382 $\pm$ 0.092
    & 0.422 $\pm$ 0.055
    & 0.745 $\pm$ 0.069
    & \textbf{0.825} $\pm$ 0.034 \\
    \midrule
    Average on all scenes 
    & 0.364 
    & 0.542 
    & 0.643 
    & \textbf{0.726} \\
    \bottomrule
  \end{tabular}
  }}
  \caption{{\bf AUCs of surface coverage on large 3D scenes.} All methods use perfect depth maps as input except for MACARONS, which takes RGB images as input. We follow \cite{guedon-nips22-scone-surface-coverage} and compute the area under the curve representing the evolution of the total surface coverage during exploration. The 8 scenes above the bar were seen by MACARONS during self-supervised training (but with different, random starting camera poses and trajectories), and the 4 scenes below the bar were not. Other methods are trained on ShapeNet~\cite{chang-15-shapenet} with 3D supervision. Even if it only uses RGB images, our model MACARONS is able to outperform the baselines in large environments since, contrary to other methods, its self-supervised online training strategy allows it to scale its learning process to any kind of environment.}
  \label{tab:coverage_auc}
\end{table*}
We compare our method to the state of the art for learning-based NBV computation for dense reconstruction in large environments. All methods use perfect depth maps as input except for MACARONS, which takes RGB images as input. We generate input data from 3D meshes of large scenes~(courtesy of Brian Trepanier and Andrea Spognetta, under CC License; all models were downloaded from the website Sketchfab). This dataset was introduced in~\cite{guedon-nips22-scone-surface-coverage}. To compare the different methods, we follow \cite{guedon-nips22-scone-surface-coverage} and compute the area under the curve of the evolution of the total surface coverage during exploration, after 100 NBV iterations, as presented in Table~\ref{tab:coverage_auc}. 
The surface coverage is computed using the ground truth meshes. AUCs are averaged on multiple trajectories in each scene: We use the same starting camera poses and same sets of candidate camera poses $C_t$ for each method for fair comparison. 
For this experiment, MACARONS was trained on a set of several scenes: The 8 scenes above the bar in Table~\ref{tab:coverage_auc} were seen during online training (but with different, random starting camera poses and trajectories), and the 4 scenes below the bar were not. The other methods were trained on ShapeNet~\cite{chang-15-shapenet} with 3D supervision since their learning process cannot scale to unknown, large environments.

During this experiment, we freeze all weights of MACARONS and only perform inference computation to better demonstrate the ability of our model to generalize to novel scenes, even when online learning is deactivated.
Even if it only uses RGB images, our model is able to outperform the baselines in large environments since, contrary to other methods, its self-supervised online training strategy allows it to scale its learning process to any kind of unknown environment, where no ground truth is available and data has to be acquired with a camera.
%
Figures~\ref{fig:supp-mat-trajectory} 
and~\ref{fig:surface-reconstruction} show examples of trajectories as well as of surface reconstructions computed with MACARONS.

\subsection{Ablation study}
\begin{table*}
  \centering
  {\scriptsize
    \addtolength{\tabcolsep}{-2pt}
  \scalebox{0.93}{

\begin{tabular}{@{}lccccccccccccccccc@{}}
    \toprule
    \multicolumn{1}{c}{} & \multicolumn{8}{c}{Categories seen during training} & \multicolumn{8}{c}{Categories not seen during training}\\
    \cmidrule(r){2-9} \cmidrule(r){10-17}
     Method & Airplane & Cabinet & Car & Chair & Lamp & Sofa & Table & Vessel & Bus & Bed & Bookshelf & Bench & Guitar & Motorbike & Skateboard & Pistol & Mean \\
    \midrule
    Random & 0.745 & 0.545 & 0.542 & 0.724 & 0.770 & 0.589 & 0.710 & 0.674 
    & 0.609 & 0.619 & 0.695 & 0.795 & 0.795 & 0.672 & 0.768 & 0.614 & 0.679 \\
    Proximity Count~\cite{delmerico-18-acomparison} & 0.800 & 0.596 & 0.591 & 0.772 & 0.803 & 0.629 & 0.753 & 0.706 
    & 0.646 & 0.645 & 0.749 & 0.829 & 0.854 & 0.705 & 0.828 & 0.660 & 0.723 \\
    Area Factor~\cite{vasquez-14-volumetric-nbv} & 0.797 & 0.585 & 0.587 & 0.751 & 0.801 & 0.627 & 0.725 & 0.714 
    & 0.629 & 0.631 & 0.742 & 0.827 & 0.852 & 0.718 & 0.799 & 0.660 & 0.715 \\
    NBV-Net~\cite{mendoza-prl2020} & 0.778 & 0.576 & 0.596 & 0.743 & 0.791 & 0.599 & 0.693 & 0.667 
    & 0.654 & 0.628 & 0.729 & 0.824 & 0.834 & 0.710 & 0.825 & 0.645 & 0.706 \\
    PC-NBV~\cite{zeng-icirs20-pcnbv} & 0.799 & 0.612 & \textbf{0.612} & \textbf{0.782} & 0.800 & 0.640 & 0.760 & 0.719 
    & 0.667 & 0.662 & 0.740 & \textbf{0.845} & 0.849 & 0.728 & 0.840 & 0.672 & 0.733 \\
    SCONE~\cite{guedon-nips22-scone-surface-coverage} & 0.827 & 0.625 & 0.591 & \textbf{0.782} & 0.819 & \textbf{0.662} & 0.792 & 0.734 
    & 0.694 & 0.689 & 0.746 & 0.832 & \textbf{0.860} & 0.728 & 0.845 & 0.717 & 0.746 \\
    MACARONS-NBV & \textbf{0.830} & \textbf{0.639} & 0.595 & 0.771 & \textbf{0.826} & \textbf{0.662} & \textbf{0.810} & \textbf{0.741} 
    & \textbf{0.702} & \textbf{0.690} & \textbf{0.758} & 0.829 & 0.852 & \textbf{0.734} & \textbf{0.848} & \textbf{0.732} & \textbf{0.751}\\
    \bottomrule
  \end{tabular}
  }
  }
  \caption{{\bf AUCs of surface coverage for several NBV selection methods for single object reconstruction, as computed on the ShapeNet test dataset following the protocol of~\cite{zeng-icirs20-pcnbv,guedon-nips22-scone-surface-coverage}.} MACARONS-NBV is trained with 3D supervision on the ShapeNet dataset using the new loss we introduced. Even if our loss is designed for large environments, it still maintains state of the art performance for the specific case of isolated, single object reconstruction.}
  \label{tab:coverage_auc_shapenet}
\end{table*}

\begin{table}
  \centering
  {\scriptsize
  \scalebox{1}{
  \begin{tabular}{@{}lccc@{}}
    \toprule
    \multicolumn{1}{c}{} & \multicolumn{2}{c}{MACARONS-NBV} & MACARONS \\
    \cmidrule(r){2-3}
     & Loss from~\cite{guedon-nips22-scone-surface-coverage} & Our loss &  \\
    \midrule
    AUC of surface coverage 
    & 0.561 $\pm$ 0.162
    & 0.635 $\pm$ 0.148
    & \textbf{0.719} $\pm$ 0.109 \\
    \bottomrule
  \end{tabular}
  }}
  \caption{{\bf Contribution of our new loss and self-supervised learning process.} 
   AUCs of surface coverage in large 3D scenes, averaged over multiple trajectories in all 12 scenes of the dataset. Both our new loss and the self-supervised learning process of the full model lead to dramatic increase in performance.}
  \label{tab:loss_ablation}
\end{table}

Apart from adapting learning-based NBV prediction to RGB inputs, we proposed both a novel loss to learn the surface coverage gain compared to~\cite{guedon-nips22-scone-surface-coverage}, and a new online training strategy to let the model learn from any kind of environment in a self-supervised fashion. To quantify the benefits of each of these two improvements, we perform two additional experiments.

First, we use our new loss function to train our volume occupancy module and surface coverage gain module for isolated, single objects with explicit 3D supervision on the ShapeNet dataset~\cite{chang-15-shapenet}, similar to \cite{guedon-nips22-scone-surface-coverage}. We call the resulting model MACARONS-NBV. We compare with other methods in Table~\ref{tab:coverage_auc_shapenet} and verify that our new loss does not lower but slightly increases performance compared to the state of the art for the specific case of single object reconstruction.

Then, we reconstruct 3D scenes with both our full model and MACARONS-NBV. For the latter, we use perfect depth maps rather than the depth prediction module. We compare two versions of MACARONS-NBV: One is trained on ShapeNet using the loss from~\cite{guedon-nips22-scone-surface-coverage}, the other is trained using our new loss. Table~\ref{tab:loss_ablation} shows how our new loss is crucial to increase performance in large scenes, and how self-supervised learning increases performance even further.

\section{Conclusion}
\label{sec:conclusion}

Our method can explore large scenes to efficiently reconstruct them using only a color camera. Beyond the potential applications, it shows that it is possible to jointly learn to explore and reconstruct a scene without any 3D input. 

We assume the scene to be static, which can be a limitation, however several self-supervised depth prediction models already showed how to be robust to moving objects~\cite{watson-cvpr21-temporal-opportunist-manydepth}. Another limitation is that we assume the pose to be known as in previous works on NBV prediction. This is reasonable as the method controls the camera but such control is never perfect. It would be interesting to estimate the camera pose as well. Since we control the camera, we already have a very good initialization, which should considerably help convergence.
We also use a very simple path planning policy from our coverage predictions, by evaluating camera poses sampled in the surroundings at each iteration. It would be very interesting to consider longer-term planning, to generate even more efficient trajectories.

\paragraph{Acknowledgements.}
This work was granted access to the HPC resources of IDRIS under the allocation 2022-AD011013387 made by GENCI. We thank Elliot Vincent for inspiring discussions and valuable feedback.


{\small

\bibliographystyle{ieee_fullname}
\bibliography{arxiv}
}

\newpage
\appendix
\section*{Appendix}
In this appendix, we provide the following elements:
\begin{enumerate}
    \item Further details about the architecture of our model MACARONS and its different modules.
    \item Further details about the training process, as well as the implementation of MACARONS.
    \item Further quantitative and qualitative results.
\end{enumerate}
We also provide on this \href{https://imagine.enpc.fr/~guedona/MACARONS/}{webpage} a video that illustrates how MACARONS explores and reconstructs large 3D structures.

\section{Architecture}

In the following subsections, we provide details about the architecture of MACARONS and its different modules.

\subsection{Depth prediction module}

\begin{figure*}
  \centering
  \includegraphics[width=.8\linewidth]{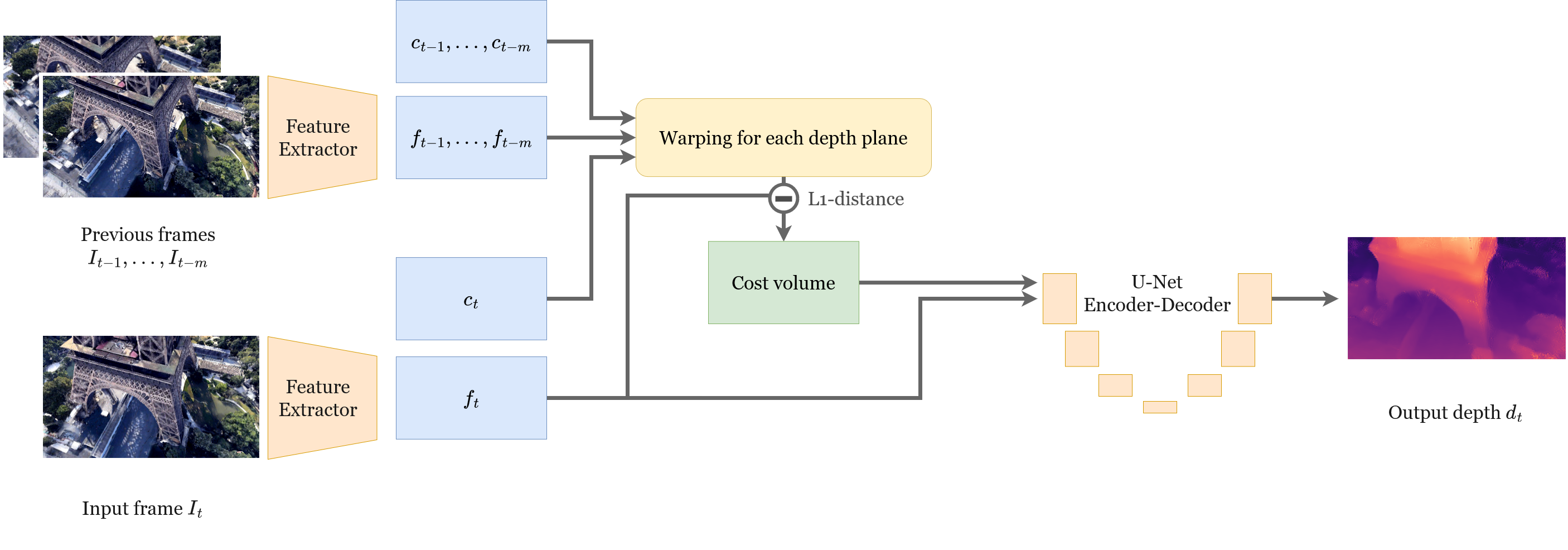}
  \caption{{\bf Architecture of the depth prediction module.} The depth prediction module relies on a cost volume to predict depth from multiple RGB inputs: Features extracted from previous images $I_{t-1}, ..., I_{t-m}$ are warped into the view space of image $I_t$ for multiple depth planes, and compared to the features extracted from $I_t$ using L1-distance.}
  \label{fig:supp-mat-depth-module}
\end{figure*}

The figure~\ref{fig:supp-mat-depth-module} illustrates the architecture of the depth prediction module of MACARONS, which takes inspiration from Watson \etal~\cite{watson-cvpr21-temporal-opportunist-manydepth}.

In our experiments, we follow~\cite{watson-cvpr21-temporal-opportunist-manydepth} and use a set of $n_{depth} = 96$ ordered planes perpendicular to the optical axis at $I_t$. The depths are linearly spaced between extremal values $d_{min}$ and $d_{max}$. We adapt $d_{min}$ and $d_{max}$ depending on the size of the bounding boxes of the scenes seen during training. We use images with size $456 \times 256$ pixels, which corresponds to a widescreen aspect ratio of 16:9.

This architecture is essential for MACARONS to learn how to compute a volume occupancy field and predict NBV in a self-supervised fashion. Indeed, we use a dense depth map prediction module rather than SfM or keypoints matching approaches because we need dense depth maps for space carving operations to generate a pseudo-GT volume occupancy and train the corresponding module.
Moreover, the depth prediction module is also a fast and precise model that allows for reconstructing in real-time the surface points seen by the camera.
%

\subsection{Volume occupancy module}

\begin{figure}
  \centering
  \includegraphics[width=0.95\linewidth]{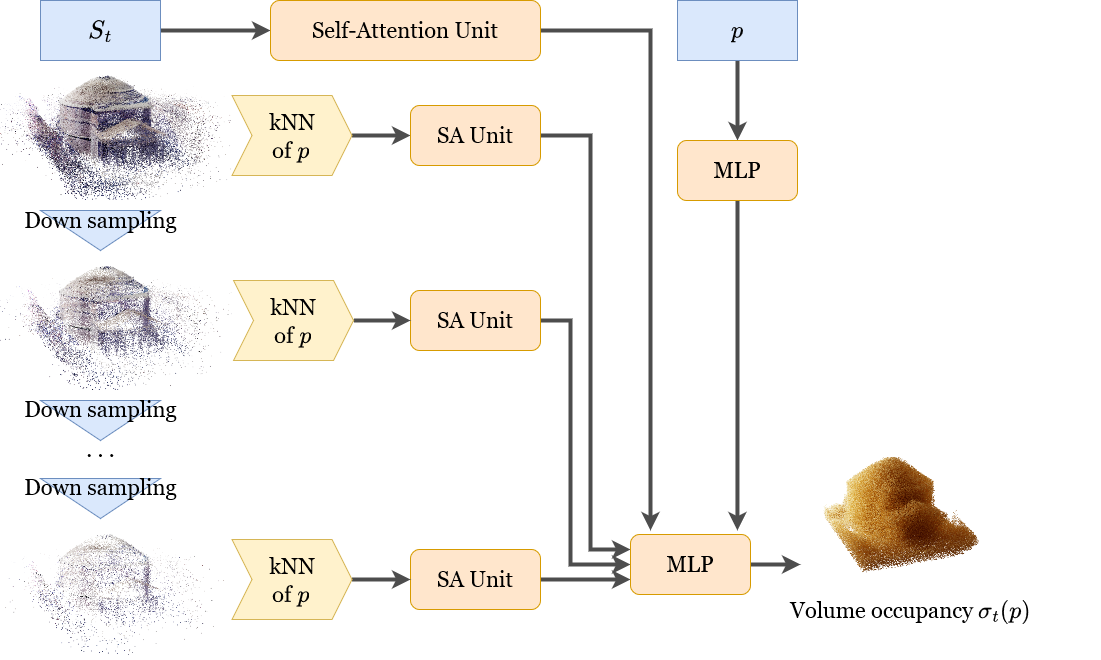}
  \caption{{\bf Architecture of the volume occupancy module.} At time step $t$, the volume occupancy module relies mostly on neighborhood features to compute its output $\sigma_t(p)$. To compute the neighborhood features of an input 3D point $p$, we apply self-attention units on the $k$-nearest neighbors of $p$ at different scales.}
  \label{fig:supp-mat-occupancy-module}
\end{figure}

The figure~\ref{fig:supp-mat-occupancy-module} presents the architecture of the volume occupancy module. We implement this module using a Transformer~\cite{vaswani-nips17-attentionisallyouneed}: The module takes as input the point $p$, the surface point cloud $S_t$ and previous poses $c_i$, and outputs a scalar value in $[0, 1]$. 

This volumetric representation is a deep implicit function inspired by~\cite{guedon-nips22-scone-surface-coverage}, and is convenient to build a NBV prediction model that scales to large environments. Indeed, as we explained in the main paper, it has a virtually infinite resolution and can handle arbitrarily large point clouds without failing to encode fine details since it uses mostly local features at different scales to compute the probability of a 3D point to be occupied.

In particular, for any 3D point $p\in\IR^3$, we compute the $k$-nearest neighbors $(q_1,...,q_k)$ of $p$ in the dense point cloud $S_t$ and transform the sequence $(q_1-p,...,q_k-p)$ using self-attention units followed by pooling operations. The resulting feature encodes information about the local state of the geometry.
Then, we iterate the process at different scales: we down-sample the point cloud, compute the $k$-nearest-neighbors, encode the sequence, and reiterate. The spatial extension of the neighbors grows as we down-sample $S_t$, which helps to encode information about geometry at larger scales.

Because this architecture relies on local, neighborhood features, it can process arbitrarily large point clouds without failing to encode fine details or producing memory issues: Indeed, for a 3D point $p$, adding distant surface points to $S_t$ does not change the local state of the geometry, nor the neighbors of $p$.

In practice, we use $k=16$ and compute features at 3 different scales.
%

\subsection{Surface coverage gain module}

\begin{figure*}
  \centering
  \includegraphics[width=\linewidth]{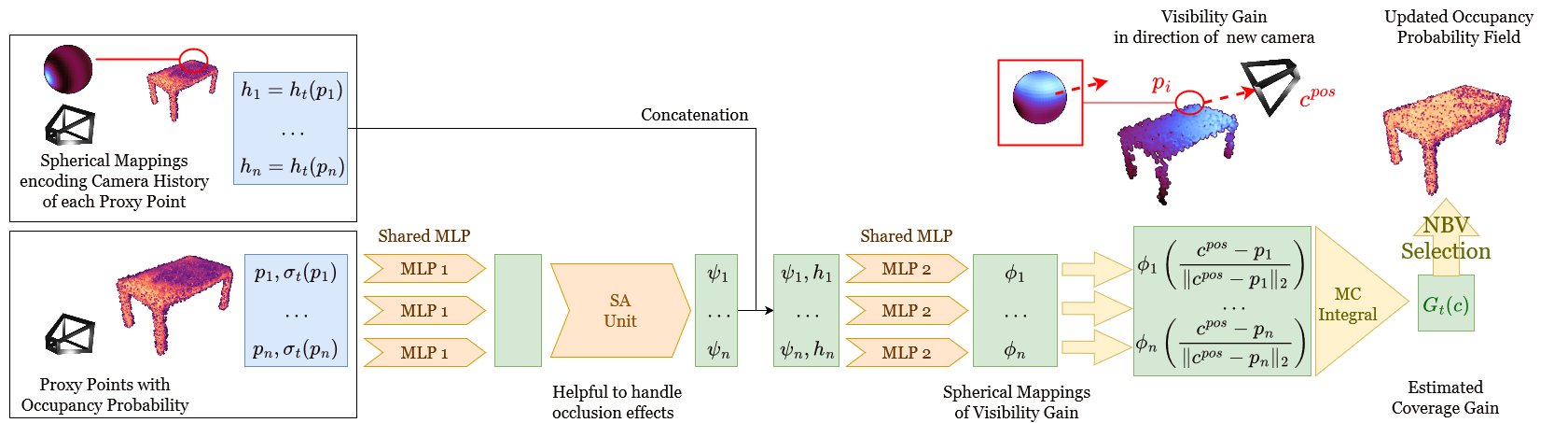}
  \caption{{\bf Architecture of the surface coverage gain module, inspired by~\cite{guedon-nips22-scone-surface-coverage}.} which predicts a visibility gain for a sequence of 3D points $p_i$ in the field of view of camera $c$. To make this prediction, the model encodes the points $p_i$ concatenated with their occupancy probability $\sigma_t(p_i)$. We use an attention mechanism to take into account occlusion effects in the volume between the 3D points and their consequences on the visibility gains. We finally use a Monte Carlo integration to compute the coverage gain $G_t(c)$ of camera $c$.}
  \label{fig:supp-mat-coverage-module}
\end{figure*}

The figure~\ref{fig:supp-mat-coverage-module} presents the architecture of the surface coverage gain module. This final module computes the surface coverage gain of a given camera pose $c$ based on the predicted occupancy field, as proposed by~\cite{guedon-nips22-scone-surface-coverage}. 
However, as we explain in the main paper, we brought key modifications to the surface coverage gain estimation formula to adapt the model to NBV computation in large environments. In the following paragraphs, we detail some technical improvements that we did not mention in the main paper.

\paragraph{Occlusion-aware camera history features.} In particular, we introduce a critical change to camera history features $h_t$. In~\cite{guedon-nips22-scone-surface-coverage}, the authors compute camera history features by projecting all previous camera positions $c_i^{pos}$ on a sphere around $p$. On the contrary, we compute $h_t(p)$ by projecting on a sphere around $p$ only the positions $c_i^{pos}$ of the cameras for which $p$ was in the field of view delimited by $c_i^{rot}$, and for which $p$ was not too far behind the surface reconstructed in the depth map $d_i$. 

Therefore, we encode only the previous camera poses for which $p$, whether it is empty or occupied, is likely to be visible. This results in camera history features that reflect previous occlusion effects. This technical detail is actually of great importance to improve performance in large environments, since there is great variability in camera fields of view.

Indeed, to compute the visibility gain of a 3D point $p$, the surface coverage gain module exploits information about the volume occupancy, the camera history and the occlusions. For the specific case of a centered object with cameras sampled on a sphere around it, the volume $\chi$ is entirely contained in the field of view of all cameras. Therefore, by encoding occlusion effects with its transformer architecture, the module is able to identify, for any camera $c_i$ in the camera history, if the point $p$ was visible from $c_i$. Consequently, projecting the positions of all previous cameras $c_i$ on a sphere around $p$ does not decrease performance since the model is able to identify which camera $p$ was visible from by encoding all occlusion effects in $\chi$.

However, in a large environment, a subset of points in $\chi$ that occludes $p$ in the direction of a previous camera $c'$ could be located outside the field of view of another new camera $c$. Thus, the visibility module could lack information about previous occlusion effects when processing the field of view of the new camera $c$ while still be provided with the information that $p$ was observed in the direction of $c'$ because of the camera history feature. This could trick the model into thinking $p$ is empty even if it is not. Consequently, modifying the camera history feature to reflect previous occlusion effects leads to better performance in large environments.

\paragraph{Additional details.} At time step $t$, all surface coverage gain predictions are made in the view space of the current camera pose $c_t$. This increases performance since it slightly simplifies the problem (on the contrary, estimating surface coverage gains from any random coordinate space, for example, would make the problem more complex).

Finally, Even if its coverage gain is known to be equal to zero, we do compute a prediction for the surface coverage gain of the current camera $c_t$ and use it to compute the loss during the online, self-supervised training. This information is actually useful to help the module set the lowest visibility gain value at zero.
%

\section{Implementation details}

\subsection{Camera management}

We follow~\cite{guedon-nips22-scone-surface-coverage} and first discretize the set of all camera poses $\calC$ in the scene on a 5D grid, that correspond to coordinates $c^{pos} = (x_c, y_c, z_c)$ of the camera as well as the elevation and azimuth to encode rotation $c^{rot}$. The number of poses depends on the dimensions of the bounding box of the scene: As we explain in the main paper, this box is an input to the algorithm, as a way for the user to tell which part of the scene should be reconstructed. We follow~\cite{guedon-nips22-scone-surface-coverage} in our experiments, and discretize the 5D grid to obtain approximately 10,000 different poses in each scene.

At each time step $t$, we define the set of possible camera poses, denoted by $\calC_t \subset \calC$, as the immediate neighbors of the current camera pose $c_t$ within the 5D grid. Specifically, these poses lie within a 6-neighborhood on the 3D position grid and a 4-neighborhood on the 2D rotation grid. We exclude any neighboring pose that shares the same position $c^{pos}_t$ as $c_t$, as the depth module requires camera movement to generate depth maps using warping operations.

\subsection{Initializing the neural modules}

We have observed that during the initial training phase of MACARONS, the volume occupancy module and surface coverage gain module sometimes exhibit instability when trained from scratch with a naive initialization. 
This instability arises from the great variability and noise in the input data, which make training challenging for self-attention units and transformer architectures. Indeed, in contrast to SCONE~\cite{guedon-nips22-scone-surface-coverage}, which has similar modules trained under ideal conditions with perfectly known 3D objects and 3D supervision, MACARONS reconstructs 3D from partial observations of unknown scenes, which results in a wide variation in batch size as well as noise in the 3D reconstructions.

However, we have found that this instability occurs only during the first few minutes of training, while both SCONE~\cite{guedon-nips22-scone-surface-coverage} and MACARONS require several dozen hours to converge. 
To stabilize the modules of MACARONS, we have developed a simple initialization process that involves a few iterations under ideal conditions: We first initialize the modules using standard techniques, such as Kaiming~\cite{he-iccv15-delvingdeepintorectifiers} for all layers except the self-attention units, for which we use Xavier initialization~\cite{glorot-10-understandingthedifficulty}. 
We then perform a few iterations with perfectly known objects, similar to~\cite{guedon-nips22-scone-surface-coverage}, which takes less than 5 minutes (1 minute is sufficient). 

We can use either a few ShapeNet~\cite{chang-15-shapenet} meshes, or simple virtual cube meshes generated online for this initialization process, eliminating the need for an additional 3D dataset.

\subsection{Training the neural modules}
To train our model in an online, self-supervised fashion for our experiments, we start by loading a 3D scene from a subset of the dataset introduced in~\cite{guedon-nips22-scone-surface-coverage}. 
We sample a random camera pose in the scene, and let our model explore. The camera captures images with size $456\times 256$ pixels, which corresponds to a widescreen aspect ratio of 16:9. The model performs NBV training iterations as described in the main paper: In particular, it builds a Memory in real-time and simultaneously learns to reconstruct surfaces and optimize its path in the volume to increase its coverage of the surface.
When starting a trajectory, we select a certain number $N$ of 3D points within the bounding box of the scene, which we refer to as \emph{proxy points}. During the online training process, we solely use these points to represent the volume: We calculate the volume occupancy exclusively for these proxy points and sample from them to predict the surface coverage gains. We save both the proxy points and their pseudo ground-truth volume occupancy values in memory. In our implementation, we typically sample 100,000 proxy points and evaluate surface coverage gains for 4 camera poses at each iteration.

After 100 NBV iterations, we load another scene and start a new trajectory. 
We perform data augmentation during training: we apply color jitter on RGB images, and perform rotations and mirroring operations on 3D inputs. 
Multi-GPU programming can be used to let the model explore several scenes at the same time and speed up convergence; In practice, we use 4 GPUs Nvidia Tesla V100 SXM2~32~Go to let the model explore 4 different scenes in parallel.

We perform up to 360 trajectories to make the model converge. However, such numbers are prone to variations: they depend not only on the complexity of the scenes but also on the size of the Memory and the number of Memory Replay iterations. 
Increasing the number of Memory Replay iterations slows down the exploration process during online training but considerably accelerates the convergence.

\subsection{Memory Replay} 

The use of Memory Replay iterations allows for training the model with more complex camera configurations, for example by evaluating the surface coverage gains of distant camera poses stored into the memory.
On the other hand, decreasing the number of memory replay iterations results in the model relying mostly on the current images for training, thus comparing surface coverage gains between nearby camera poses.

In our experiments, for each GPU, we store the data from the last 10 trajectories into the memory. We use 5 memory replay iterations for the depth module and only 1 for both volume occupancy and surface coverage gain modules, by using the 4 latest images captured from nearby camera poses. The self-supervision signal is built by comparing the current state of the scene to the state of the same scene before capturing these images.
The use of a single memory replay iteration for the volume occupancy and surface coverage gain modules results in good performance since we select the next best view from nearby camera poses in our naive path planning strategy.

However, more complex strategies for selecting the Next Best View (NBV), which include distant camera poses, could be devised. In such cases, increasing the number of memory replay iterations to create a self-supervision signal that aligns with the expected camera configuration should improve the performance.
%

\subsection{Computational cost}

Generally, computing a whole trajectory only takes a few minutes at inference, even on a single GPU Nvidia GeForce GTX 1080~Ti. During online, self-supervised training, computing a whole trajectory can take up to 10 minutes using GPU Nvidia Tesla V100 in our main experimental setup (which consists in 5 memory replay iterations for depth module and only 1 for volume occupancy and surface coverage gain modules), and up to 25 minutes for a different setup (5 memory replay iterations for depth module and 3 for the other modules).

However, we train MACARONS in synthetic scenes, which requires the GPU to perform numerous rendering operations to produce RGB inputs. As a result, the model's processing speed should be much faster in real-world scenarios. Specifically, with online learning activated, MACARONS can process 1.33 frames per second. After sufficient training, online learning can be disabled even in new, unfamiliar scenes, increasing the processing rate to 2.35 frames per second.
These processing rates make MACARONS well-suited for real-time exploration, as it is not necessary to process every frame captured by the camera, but only a small subset of them.

\section{Memory Building}

\subsection{Partial surface point cloud}

As we explained in the main paper, to compute the reconstructed surface point cloud $S_t$ at time step $t$ we backproject the depth map $d_t$ in 3D, filter the point cloud and concatenate it to the previous points obtained from $d_0, .., d_{t-1}$.
We filter points associated to strong gradients in the depth map, which we observed are likely to yield wrong 3D points: We remove points based on their value for the edge-aware smoothness loss appearing in~\cite{watson-cvpr21-temporal-opportunist-manydepth,godard-cvpr17-unsupervisedmonoculardepth,heise-iccv13-pmhuber} that we also use for training.
We hypothesize such outliers are linked to the module incapacity to output sudden changes in depth, thus resulting in over-smooth depth maps.

To avoid processing an excessively large point cloud, we choose to backproject only a randomly selected subset of the $456 \times 256$ pixels contained in the depth map. In the experiments we conducted, we sampled $5\%$ of the pixels, resulting in the backprojection of 5836 pixels for each new depth map produced by the model.

\subsection{Pseudo-GT volume occupancy}

We rely on Space Carving~\cite{kutulakos00space} using the predicted depth maps to create a supervision signal to train the prediction of the volume occupancy field. As explained in the main paper, our key idea is as follows: When the whole surface of the scene is covered with depth maps, a 3D point $p \in \IR^3$ is occupied iff for any depth map $d$ containing $p$ in its field of view, $p$ is located behind the surface visible in $d$.

In practice, at time step $t$, we only have access to the depth maps $d_{t, 1}', ..., d_{t,n}'$ predicted for the images captured so far. We can still compute an intermediate occupancy field, which is an approximation but can be used as supervision signal. Since it is not reliable far away from the depth maps when the whole surface has not been covered, we only sample points around the newly reconstructed surface within a margin that increases with the total number of depth maps. In our experiments, this margin increases during the trajectory from 0 to approximately half the length of the bounding box, depending on the scene. We use the $\arctan$ function to compute the margin, rather than using a linear growth.

Finally, the depth maps generated by our model are not perfect, and some of them could contain errors. Therefore, eliminating all proxy points that are not located behind depth maps can be too aggressive and may produce inaccurate volume occupancy fields. To alleviate this harsh space carving approach, we introduce two simple ideas: a \emph{score-based carving operation} and a \emph{carving tolerance}.

\paragraph{Score-based carving.} We assume that only a small portion of the depth maps produced by our model are inaccurate. Thus, we suggest to keep proxy points iff they are located behind a sufficient number of depth maps. 
Specifically, for a proxy point $p$, we denote by $n_d(p)$ the number of depth maps for which $p$ is the field of view of the depth map, and we denote by $n_b(p)$ the number of depth maps for which $p$ is not only in the field of view of the depth map, but also located behind the surface shown in the depth map.
We finally define the score of $p$ as $s(p) = \frac{n_b(p)}{n_d(p)}$. To perform space carving, we preserve proxy points with a score exceeding a certain threshold. We set this threshold to 0.95 in our experiments.

\paragraph{Carving tolerance.} To further alleviate space carving, we allow 3D points to be located \emph{in front of} a depth map but only up to a certain distance that depends on the size of the bounding box of scene. We refer to this range as the \emph{carving tolerance}, and we set it to roughly $5\%$ of the spatial extent of the bounding box in our experiments.

\subsection{Pseudo-GT surface coverage gain}

When computing pseudo-GT surface coverage gains, we count the number of new visible surface points in newly acquired depth maps. However, surface points have to be uniformly sampled on the whole surface to allow for accurate supervision coverage gain computation. Indeed, if surface points in $S_t$ are not uniformly sampled on the surface, then the pseudo-GT coverage gain will be higher than expected in areas where the surface points are the most concentrated.

To address this issue, we apply a filtering process to the surface point cloud $S_t$ during online training: we regularly recompute a filtered version $S'_t$ of $S_t$ by redistributing the surface points in small cells across the volume, which contain approximately the same number of surface points. In our experiments, we typically use 50 to 150 cells depending on the spatial extent of the scene, and set the maximum capacity of a cell to 1,000 points.
This simple approach asymptotically promotes the uniform distribution of points on the surface. We recompute $S'_t$ every 20 training iterations and fill the cells incrementally by adding 1,000 points at a time. Indeed, we encountered some issues when filling cells with a large number of points at the same time; Therefore, filling incrementally the cells gives better performance.

\section{Experiments}

In this section, we first provide extensive details concerning the experiment presented in subsection 5.2. of the main paper. Then, we present additional analysis about the benefits brought by our approach.

\subsection{Exploration of large 3D scenes}

We first provide further analysis concerning the main experiment presented in the paper, which compares our approach MACARONS to different NBV baselines for automated exploration and reconstruction of large 3D scenes. 

\begin{figure*}
  \centering
  \begin{subfigure}{0.23\linewidth}
    \includegraphics[width=\linewidth]{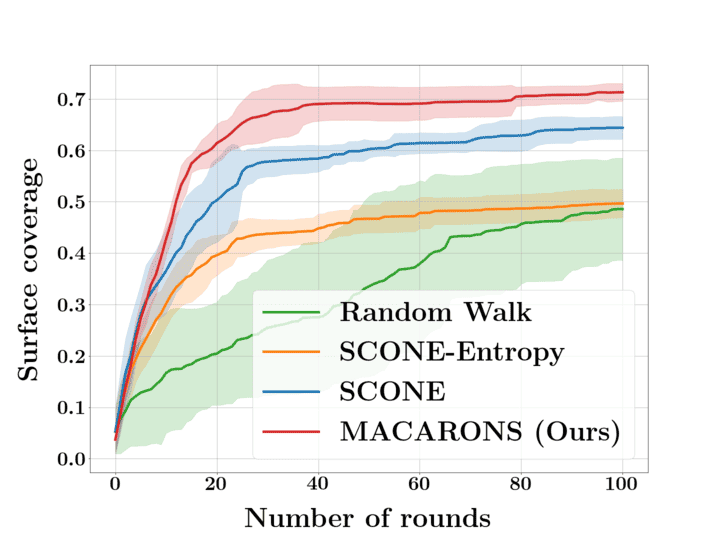}
    \caption{Pisa Cathedral}
    \label{fig:coverage-a}
  \end{subfigure}
  \hfill
  \begin{subfigure}{0.23\linewidth}
    \includegraphics[width=\linewidth]{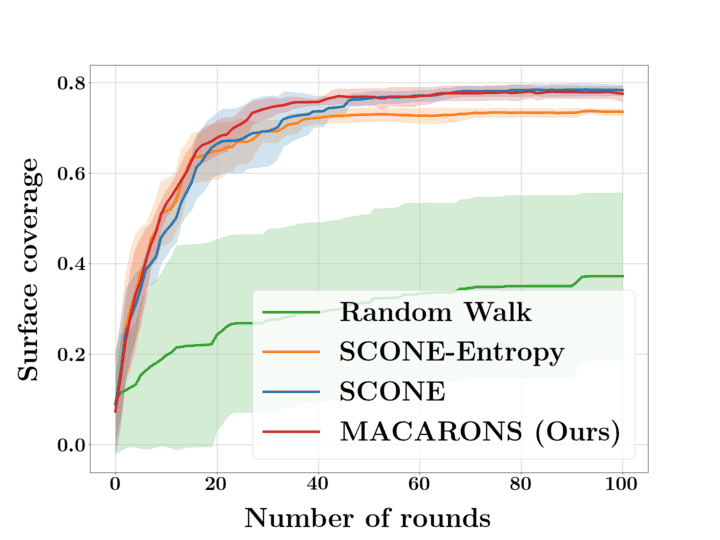}
    \caption{Statue of Liberty}
    \label{fig:coverage-b}
  \end{subfigure}
  \hfill
  \begin{subfigure}{0.23\linewidth}
    \includegraphics[width=\linewidth]{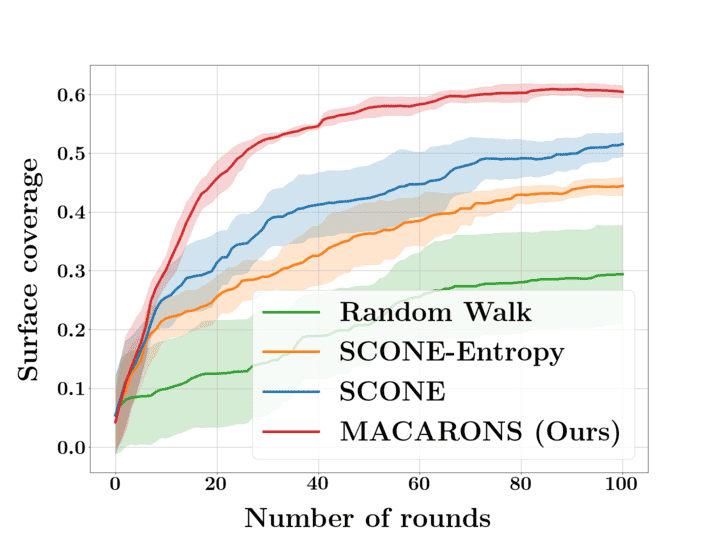}
    \caption{Pantheon}
    \label{fig:coverage-c}
  \end{subfigure}
  \hfill
  \begin{subfigure}{0.23\linewidth}
    \includegraphics[width=\linewidth]{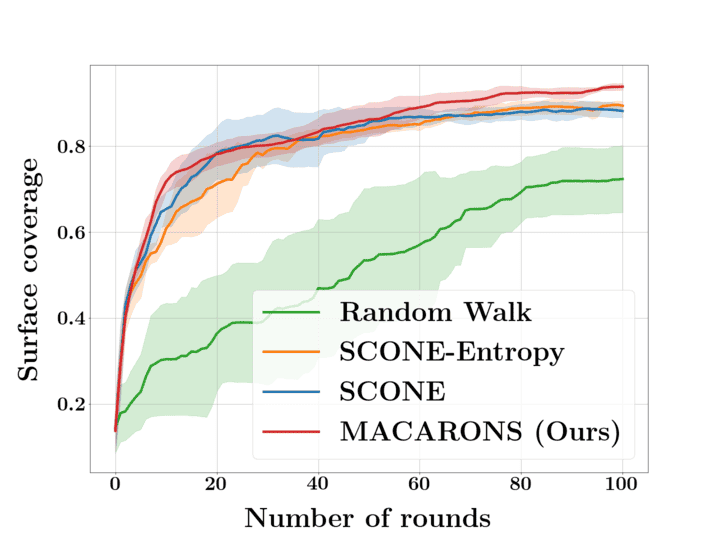}
    \caption{Fushimi Castle}
    \label{fig:coverage-d}
  \end{subfigure} \\
  \begin{subfigure}{0.23\linewidth}
    \includegraphics[width=\linewidth]{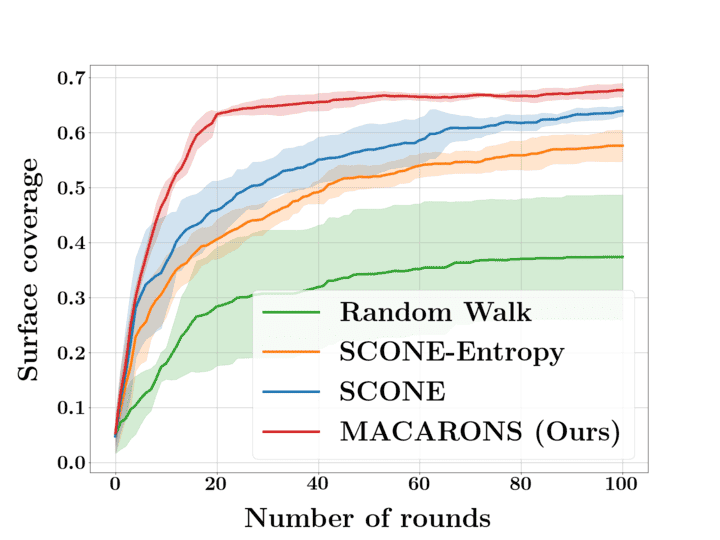}
    \caption{Colosseum}
    \label{fig:coverage-e}
  \end{subfigure}
  \hfill
  \begin{subfigure}{0.23\linewidth}
    \includegraphics[width=\linewidth]{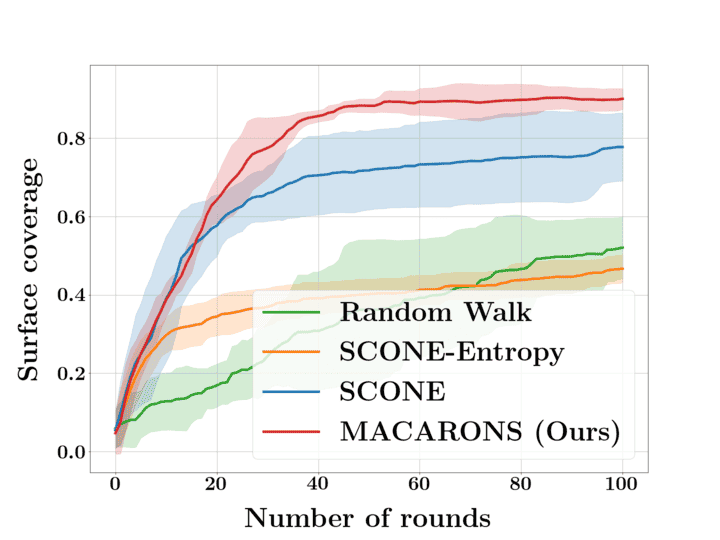}
    \caption{Dunnottar Castle}
    \label{fig:coverage-f}
  \end{subfigure}
  \hfill
  \begin{subfigure}{0.23\linewidth}
    \includegraphics[width=\linewidth]{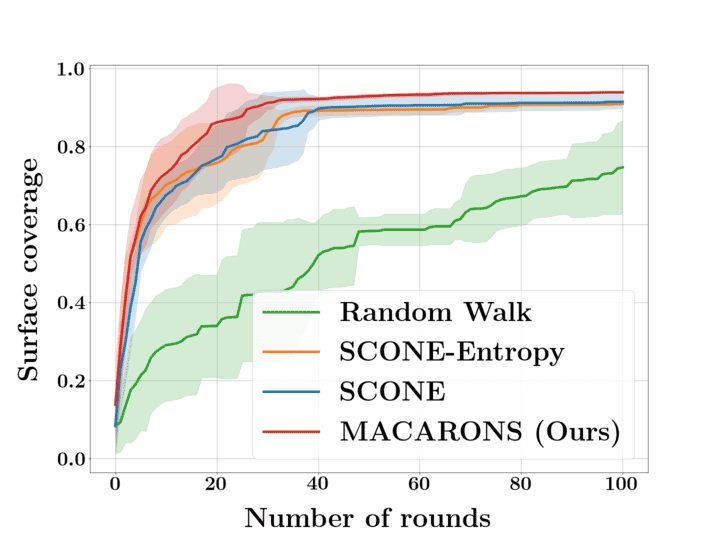}
    \caption{Christ the Redeemer}
    \label{fig:coverage-g}
  \end{subfigure}
  \hfill
  \begin{subfigure}{0.23\linewidth}
    \includegraphics[width=\linewidth]{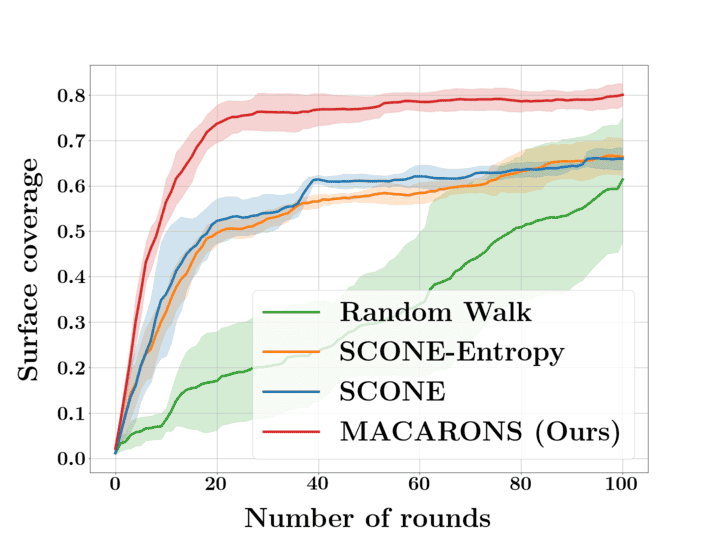}
    \caption{Bannerman Castle}
    \label{fig:coverage-h}
  \end{subfigure}
  \begin{subfigure}{0.23\linewidth}
    \includegraphics[width=\linewidth]{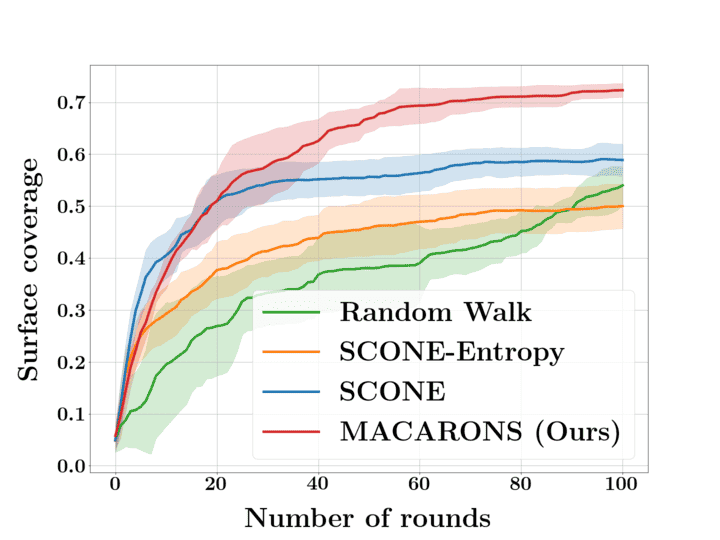}
    \caption{Alhambra Palace}
    \label{fig:coverage-i}
  \end{subfigure}
  \hfill
  \begin{subfigure}{0.23\linewidth}
    \includegraphics[width=\linewidth]{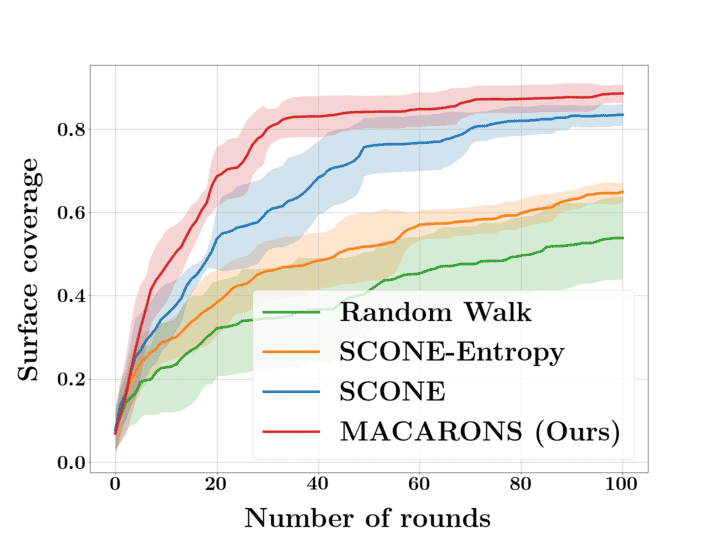}
    \caption{Neuschwanstein Castle}
    \label{fig:coverage-j}
  \end{subfigure}
  \hfill
  \begin{subfigure}{0.23\linewidth}
    \includegraphics[width=\linewidth]{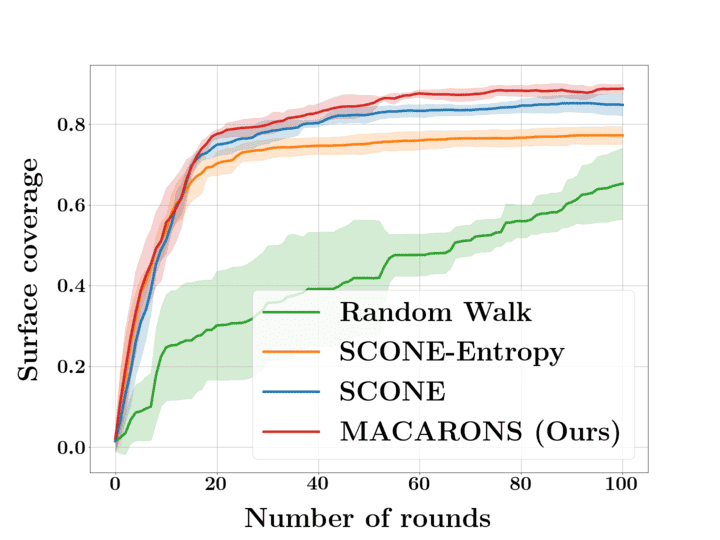}
    \caption{Eiffel Tower}
    \label{fig:coverage-k}
  \end{subfigure}
  \hfill
  \begin{subfigure}{0.23\linewidth}
\includegraphics[width=\linewidth]{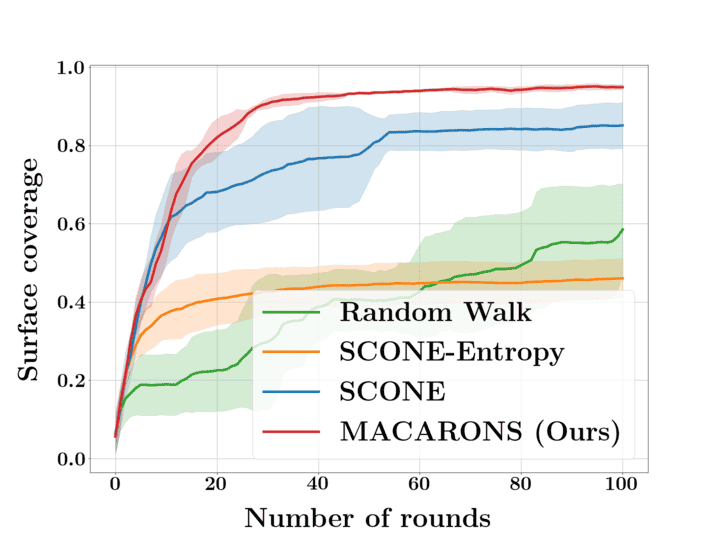}
    \caption{Manhattan Bridge}
    \label{fig:coverage-l}
  \end{subfigure} \\
  \begin{subfigure}{0.23\linewidth}
\includegraphics[width=\linewidth]{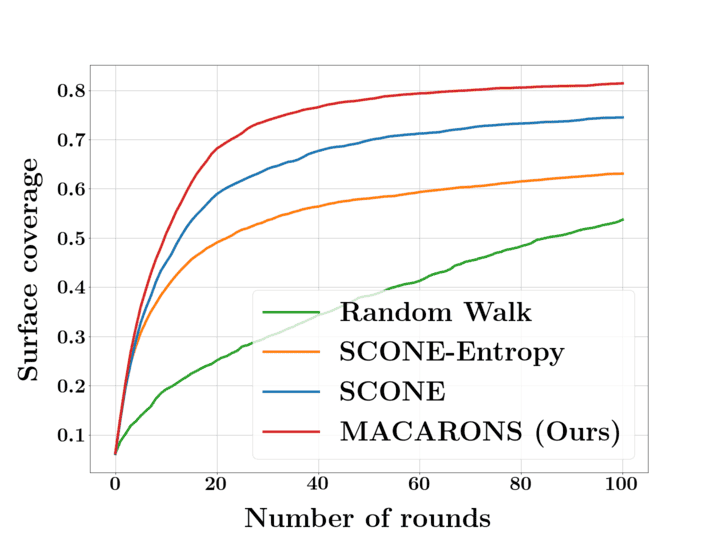}
    \caption{Average on all scenes}
    \label{fig:coverage-m}
  \end{subfigure}
  \caption{{\bf Convergence speed of the surface coverage in large 3D scenes by MACARONS and several baselines from \cite{guedon-nips22-scone-surface-coverage}.} All methods use perfect depth maps as input except for MACARONS, which takes RGB images as input. We follow \cite{guedon-nips22-scone-surface-coverage} and plot the evolution of the total surface coverage during exploration, after 100 NBV iterations. We average surface coverage on several trajectories for each scene, starting from random camera poses. Standard deviations are shown on the figures. Our model MACARONS has been trained on a set of previous scenes; all weights are frozen and we only perform inference computation. Other methods are trained on ShapeNet~\cite{chang-15-shapenet} with 3D supervision. The first two rows depict scenes that were already seen by MACARONS during its online, self-supervised training. The third row depicts scenes the model has never seen before. Even if it only uses RGB images, our model MACARONS is able to outperform the baselines in large environments since, contrary to other methods, its self-supervised online training strategy allows it to scale its learning process to any kind of environment.
  }
  \label{fig:coverage_evolution}
\end{figure*}

In particular, to complement the quantitative results presented in Table~1 of the main paper, we provide in Figure~\ref{fig:coverage_evolution} details about the convergence speed of the surface coverage in large 3D scenes for MACARONS and the baselines from~\cite{guedon-nips22-scone-surface-coverage}.
%

%
%

Finally, we provide on \href{https://imagine.enpc.fr/~guedona/MACARONS/}{this website} a video that illustrates how MACARONS explores and reconstructs efficiently a subset of three large 3D scenes. In particular, the video shows several key-elements of our approach:
\begin{enumerate}
    \item The trajectory of the camera, evolving in real-time with each NBV iteration performed by the surface coverage gain module (left).
    \item The RGB input captured by the camera (top right).
    \item The surface point cloud reconstructed using the depth prediction module of MACARONS (right).
    \item The volume occupancy field computed and updated in real-time using the volume occupancy module (bottom right). In the video, we removed the points with an occupancy lower or equal to 0.5 for clarity.
\end{enumerate}

\subsection{Ablation study: Loss function}
%

We illustrate how the novel loss we designed for surface coverage gain estimation is grounded in the theoretical framework introduced in~\cite{guedon-nips22-scone-surface-coverage}. 

\paragraph{Improving surface coverage gain estimation.} As we discussed in the main paper, the formalism introduced in~\cite{guedon-nips22-scone-surface-coverage} aims to estimate the surface coverage gain by integrating over the volume occupancy. However, this estimation can only be performed to a scale factor that cannot be computed in closed form. 

\begin{figure}
  \centering

  \begin{subfigure}{0.46\linewidth}
    \includegraphics[width=\linewidth]{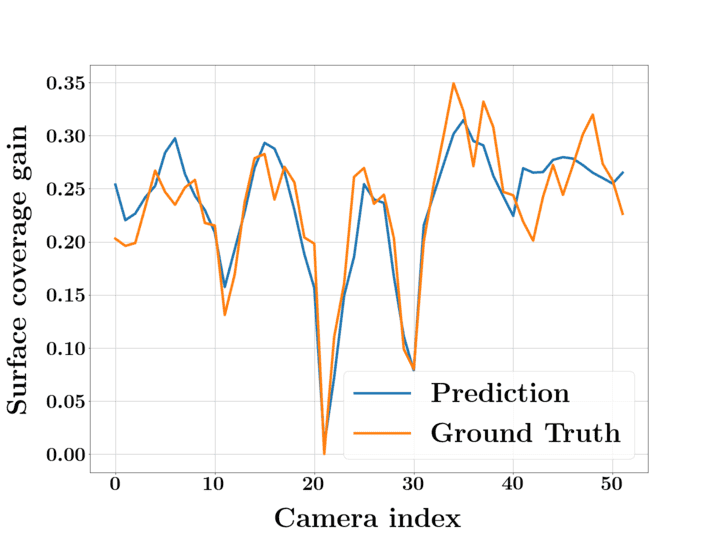}
  \end{subfigure}
  \hfill
  \begin{subfigure}{0.46\linewidth}
    \includegraphics[width=\linewidth]{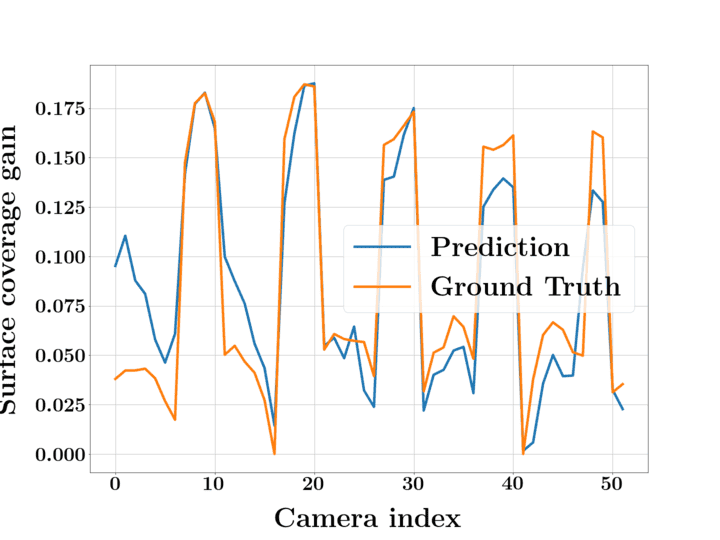}
  \end{subfigure}\\
  \begin{subfigure}{0.46\linewidth}
    \includegraphics[width=\linewidth]{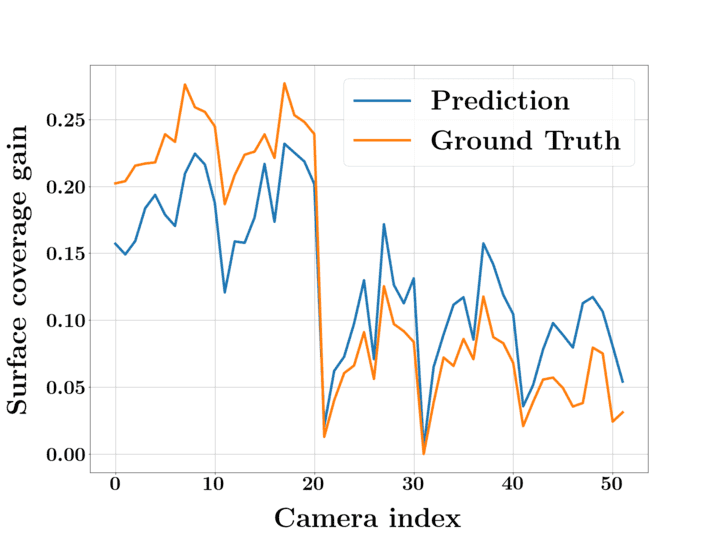}
  \end{subfigure}
  \hfill
    \begin{subfigure}{0.46\linewidth}
    \includegraphics[width=\linewidth]{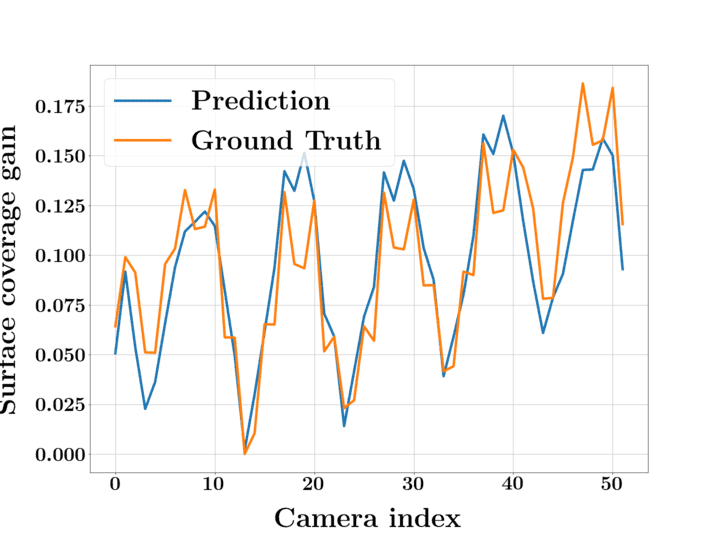}
  \end{subfigure}
  \caption{\textbf{Comparison between true surface coverage gains and predicted surface coverage gains on ShapeNet models, using our novel loss.} As expected, the normalized true and predicted coverages have great similarity, which verifies the hypothesis about the proportionality of true surface coverage gains and the predicted volumetric integrals.}
  \label{fig:supp-mat-coverage-proportionality}
\end{figure}

Consequently, the predicted surface coverage gains are supposed to be proportional to the real values. We build our novel loss on this single hypothesis: If the values are proportional, then dividing both predicted and pseudo-GT coverage gains by their respective means should result in similar values, that we directly compare with a L1-norm during training.
As shown in Figure~\ref{fig:supp-mat-coverage-proportionality}, we verify this theoretical proportionality on different samples of the ShapeNet dataset~\cite{chang-15-shapenet}. To this end, we use a version of MACARONS, called MACARONS-NBV, trained on ShapeNet only with perfect depth maps. We sample random meshes in the dataset as well as random initial camera poses; Then, we apply the volume occupancy and the surface coverage gain modules to predict the surface coverage gains of cameras sampled on a sphere around the object. We divide predicted coverage gains and ground truth coverage gains by their respective means and compare the two distributions. As expected, the two are highly similar for many objects in the dataset.

\paragraph{Additional details.} In the main paper, we compare in Table~1 our self-supervised method MACARONS trained with our novel loss to the original pipeline SCONE~\cite{guedon-nips22-scone-surface-coverage} trained on ShapeNet with the loss from~\cite{guedon-nips22-scone-surface-coverage}. We did not evaluate our new online, self-supervised pipeline with the loss from~\cite{guedon-nips22-scone-surface-coverage} because this loss needs a dense set of cameras, which is not available during online exploration. Indeed, this loss gives chaotic results when trained with a sparse set of cameras in our pipeline.
However, we proposed in Table~3 of the main paper an ablation in large environments, that uses the same MACARONS-NBV, trained on ShapeNet only with both our loss and the loss from~\cite{guedon-nips22-scone-surface-coverage}. MACARONS-NBV performs slightly worse than SCONE~\cite{guedon-nips22-scone-surface-coverage} when trained on ShapeNet with the loss from~\cite{guedon-nips22-scone-surface-coverage}, because authors from~\cite{guedon-nips22-scone-surface-coverage} use additional, hand-crafted operations to help their model process large and unknown scenes. On the contrary, we do not use such post-processing tricks but let our model learn to process large scenes by itself with our online self-supervised training, which explains the superiority of the full model MACARONS.

\subsection{Ablation study: Pretraining, Memory Replay}

We conducted an additional ablation study in Table~\ref{tab:ablation_coverage_auc} to assess the influence of memory replay iterations and pretraining with explicit 3D supervision on~\cite{chang-15-shapenet}. We trained MACARONS using different setups: we initialize the model using our previously described initialization process (\emph{Initialized}), or with a complete pretraining on ShapeNet with explicit 3D supervision, following~\cite{guedon-nips22-scone-surface-coverage} (\emph{Pretrained}). Additionally, we perform either one memory replay iteration (\emph{1 MRI}) or three memory replay iterations (\emph{3 MRI}) for both the volume occupancy and surface coverage gain modules. For the depth module, we computed five memory replay iterations in each setup.

As previously mentioned, the model does not gain from additional memory replay iterations when trained from scratch because it prioritizes learning to predict surface coverage gains for nearby camera poses due to our unsophisticated path planning strategy. However, if the model has already acquired knowledge about NBV prediction through a full pretraining program involving explicit 3D supervision on ShapeNet, it appears to benefit from more memory replay iterations as they allow for further specializing in NBV prediction in large, unknown and complex scenes.

Finally, as stated in the main paper, a basic pretraining approach on ShapeNet with explicit 3D supervision, without incorporating a self-supervision strategy in unfamiliar environments, fails in achieving performance comparable to the full model MACARONS. Indeed, even with additional hand-crafted tricks to adapt NBV-prediction to larger scenes, SCONE's performance is still inferior to MACARONS when trained from scratch with self-supervision only.

\begin{table*}
  \centering
  {
  \scalebox{0.8}{
  \begin{tabular}{@{}lccccc@{}}
    \toprule
    \multicolumn{1}{c}{} & \multicolumn{1}{c}{} & \multicolumn{4}{c}{MACARONS} \\
    \cmidrule(r){3-6}
     & & Initialized & Initialized & Pretrained & Pretrained \\
    3D scene & SCONE~\cite{guedon-nips22-scone-surface-coverage} & 1 MRI & 3 MRI & 1 MRI & 3 MRI \\
    \midrule 
Dunnottar Castle 
& 0.650 $\pm$ 0.093
& 0.774 $\pm$ 0.049
& 0.791 $\pm$ 0.049
& 0.784 $\pm$ 0.025
& \textbf{0.811} $\pm$ 0.025 \\ 
Colosseum 
& 0.532 $\pm$ 0.032
& 0.622 $\pm$ 0.005
& 0.622 $\pm$ 0.012
& \textbf{0.629} $\pm$ 0.011
& 0.620 $\pm$ 0.011 \\ 
Bannerman Castle 
& 0.552 $\pm$ 0.020
& 0.764 $\pm$ 0.026
& \textbf{0.775} $\pm$ 0.022
& 0.716 $\pm$ 0.029
& 0.763 $\pm$ 0.017 \\ 
Pantheon 
& 0.401 $\pm$ 0.030
& 0.488 $\pm$ 0.012
& 0.505 $\pm$ 0.007
& 0.499 $\pm$ 0.020
& \textbf{0.503} $\pm$ 0.014 \\ 
Christ the Redeemer 
& 0.833 $\pm$ 0.037
& 0.848 $\pm$ 0.024
& 0.833 $\pm$ 0.037
& 0.865 $\pm$ 0.037
& \textbf{0.868} $\pm$ 0.018 \\ 
Statue of Liberty 
& 0.695 $\pm$ 0.020
& 0.696 $\pm$ 0.010
& 0.670 $\pm$ 0.009
& \textbf{0.708} $\pm$ 0.014
& 0.703 $\pm$ 0.015 \\ 
Pisa Cathedral 
& 0.555 $\pm$ 0.033
& 0.633 $\pm$ 0.023
& 0.645 $\pm$ 0.008
& \textbf{0.661} $\pm$ 0.008
& 0.640 $\pm$ 0.011 \\ 
Fushimi Castle 
& 0.806 $\pm$ 0.029
& 0.807 $\pm$ 0.014
& 0.821 $\pm$ 0.013
& 0.842 $\pm$ 0.019
& \textbf{0.854} $\pm$ 0.012 \\ 
\midrule 
Alhambra Palace 
& 0.528 $\pm$ 0.030
& 0.632 $\pm$ 0.017
& 0.642 $\pm$ 0.023
& 0.634 $\pm$ 0.019
& \textbf{0.656} $\pm$ 0.011 \\ 
Neuschwanstein Castle 
& 0.662 $\pm$ 0.041
& 0.738 $\pm$ 0.043
& 0.761 $\pm$ 0.029
& 0.760 $\pm$ 0.020
& \textbf{0.766} $\pm$ 0.030 \\ 
Eiffel Tower 
& 0.753 $\pm$ 0.013
& 0.766 $\pm$ 0.033
& 0.766 $\pm$ 0.016
& 0.789 $\pm$ 0.010
& \textbf{0.794} $\pm$ 0.018 \\ 
Manhattan Bridge 
& 0.745 $\pm$ 0.069
& 0.820 $\pm$ 0.021
& 0.768 $\pm$ 0.106
& 0.825 $\pm$ 0.034
& \textbf{0.833} $\pm$ 0.021 \\ 
\midrule 
Average on all scenes 
& 0.643 
& 0.716 
& 0.717 
& 0.726 
& \textbf{0.734} \\ 
    \bottomrule
  \end{tabular}
  }}
  \caption{{\bf AUCs of surface coverage on large 3D scenes for SCONE~\cite{guedon-nips22-scone-surface-coverage} and different training configurations of MACARONS.} SCONE~\cite{guedon-nips22-scone-surface-coverage} uses perfect depth maps as input, and MACARONS takes RGB images as input. We follow \cite{guedon-nips22-scone-surface-coverage} and compute the area under the curve representing the evolution of the total surface coverage during exploration. The 8 scenes above the bar were seen by MACARONS during self-supervised training (but with different, random starting camera poses and trajectories), and the 4 scenes below the bar were not. SCONE~\cite{guedon-nips22-scone-surface-coverage} is trained on ShapeNet~\cite{chang-15-shapenet} with 3D supervision. We trained MACARONS using different setups: we initialize the model using our previously described initialization process (\emph{Initialized}), or with a complete pretraining on ShapeNet with explicit 3D supervision, following~\cite{guedon-nips22-scone-surface-coverage} (\emph{Pretrained}). Additionally, we perform either one memory replay iteration (\emph{1 MRI}) or three memory replay iterations (\emph{3 MRI}) for both the volume occupancy and surface coverage gain modules. For the depth module, we computed five memory replay iterations in each setup.}
  \label{tab:ablation_coverage_auc}
\end{table*}
\end{document}